\documentclass{article}

\PassOptionsToPackage{numbers, compress}{natbib}

\usepackage[final]{neurips_2025}

\usepackage[utf8]{inputenc} 
\usepackage[T1]{fontenc}    
\usepackage[hidelinks]{hyperref}       
\usepackage{url}
\usepackage{placeins}
\usepackage{graphicx}
\usepackage{booktabs}
\usepackage{amsmath}
\usepackage{amssymb}
\usepackage{amsfonts}
\usepackage{nicefrac}
\usepackage{microtype}
\usepackage{xcolor}
\usepackage{svg}
\usepackage{comment}
\usepackage{adjustbox}
\usepackage{tabularx}
\usepackage[inline]{enumitem}
\usepackage{cprotect}
\usepackage[noorphans,vskip=0.5ex]{quoting}
\usepackage{longtable}
\usepackage{makecell} 
\usepackage{pifont}
\usepackage{multirow}
\usepackage{wrapfig}
\usepackage{ragged2e}
\usepackage{siunitx}
\usepackage{subcaption}
\usepackage{tcolorbox}

\usepackage[symbol]{footmisc}
\renewcommand{\thefootnote}{\fnsymbol{footnote}}

\usepackage[capitalize]{cleveref}
\crefname{section}{Sec.}{Secs.}
\Crefname{section}{Section}{Sections}
\Crefname{table}{Table}{Tables}
\crefname{table}{Tab.}{Tabs.}

\newcommand{\todo}[1]{\textcolor[rgb]{1, 0, 0}{\textbf{(TODO: #1)}}}

\newcommand{\myparagraph}[1]{\vspace{3pt}\noindent\textbf{#1}}

\newcolumntype{Y}{>{\centering\arraybackslash}X}

\definecolor{turquoise}{cmyk}{0.65,0,0.1,0.3}
\definecolor{purple}{rgb}{0.65,0,0.65}

\usepackage{array}
\newcolumntype{F}[1]{%
    >{\raggedright\arraybackslash\hspace{0pt}}p{#1}}%
\newcolumntype{T}[1]{%
    >{\centering\arraybackslash\hspace{0pt}}p{#1}}

\newcommand{\eg}{\textit{e.g.}} 
\newcommand{\ie}{\textit{i.e.}} 

\definecolor{fixmecolor}{rgb}{0.1,0.1,0.9} 
\definecolor{notecolor}{rgb}{0.1,0.7,0.1} 

\newcommand{\benchmark}{{Qualcomm Interactive Cooking}}
\newcommand{\model}{\textsc{LiveMamba}}

\newcolumntype{L}{>{\RaggedRight\arraybackslash}X}

\newcommand{\commentson}{} 

\ifdefined\commentson
  \newcommand{\fixme}[1]{\textcolor{fixmecolor}{\textbf{FIXME:} #1}}
  \newcommand{\note}[1]{\textcolor{notecolor}{\textbf{NOTE:} #1}}
  \newcommand{\authcomment}[3][]{
    \textcolor{#2}{\textbf{[#1]} #3}%
  }
  \newcommand{\authA}[1]{\authcomment[AA]{orange}{#1}} 
  \newcommand{\authB}[1]{\authcomment[BB]{cyan}{#1}}   
\else 
  \newcommand{\todo}[1]{}
  \newcommand{\fixme}[1]{}
  \newcommand{\note}[1]{}
  \newcommand{\authcomment}[3][] {}
  \newcommand{\authA}[1]{}
  \newcommand{\authB}[1]{}
\fi

\usepackage[symbol]{footmisc}
\renewcommand{\thefootnote}{\fnsymbol{footnote}}
\newcommand{\anumfootnote}[1]{%
  \leavevmode\unskip
  {\renewcommand{\thefootnote}{\arabic{footnote}}%
   \footnote{#1}}%
  \ignorespaces
}

\title{Can Multi-Modal LLMs Provide Live Step-by-Step Task Guidance?}

\author{%
    Apratim Bhattacharyya\textsuperscript{1}\thanks{Equal contribution.}\hspace{0.35cm} 
    Bicheng Xu\textsuperscript{2$*$}\thanks{Work done while employed at Qualcomm AI Research.}\hspace{0.35cm}
    Sanjay Haresh\textsuperscript{1}\hspace{0.35cm}
    Reza Pourreza\textsuperscript{1}\hspace{0.35cm} \\
    \textbf{Litian Liu\textsuperscript{1}}\hspace{0.35cm}
    \textbf{Sunny Panchal\textsuperscript{1}}\hspace{0.35cm}
    \textbf{Pulkit Madan\textsuperscript{1}}\hspace{0.35cm}
    \textbf{Leonid Sigal\textsuperscript{2}}\hspace{0.35cm}
    \textbf{Roland Memisevic\textsuperscript{1}} \\
    \textsuperscript{1} Qualcomm AI Research\thanks{Qualcomm AI Research is an initiative of Qualcomm Technologies, Inc.}\hspace{0.35cm} 
    \textsuperscript{2} University of British Columbia
}

\begin{document}

\maketitle

\begin{abstract}
  Multi-modal Large Language Models (LLM) have advanced conversational abilities but struggle with providing live, interactive step-by-step guidance, a key capability for future AI assistants. 
  Effective guidance requires not only delivering instructions but also detecting their successful execution, 
  as well as identifying and alerting users to mistakes, all of which has to happen in real-time. 
  This requires models that are not turn-based, but that can react asynchronously to a video stream, 
  as well as video data showing users performing tasks including mistakes and their corrections. 
  To this end, we introduce \benchmark{}, a new benchmark and dataset built upon CaptainCook4D, which contains user mistakes during task execution. Our dataset and benchmark features densely annotated, timed instructions and feedback messages, specifically including mistake alerts precisely timestamped to their visual occurrence in the video. 
  We evaluate state-of-the-art multi-modal LLMs on the \benchmark{} benchmark and introduce \model{}, a streaming multi-modal LLM designed for interactive instructional guidance. This work provides the first dedicated benchmark and a strong baseline for developing and evaluating on live, situated coaching.
\end{abstract}

\section{Introduction}
 Multi-modal Large Language Models (LLMs) have recently advanced, enabling AI systems to interact with users more naturally, fluently, and in real-time by processing audio, speech, and visual inputs for conversations about images or videos. 

\begin{figure}[t]
    \centering
    \includegraphics[width=\textwidth]{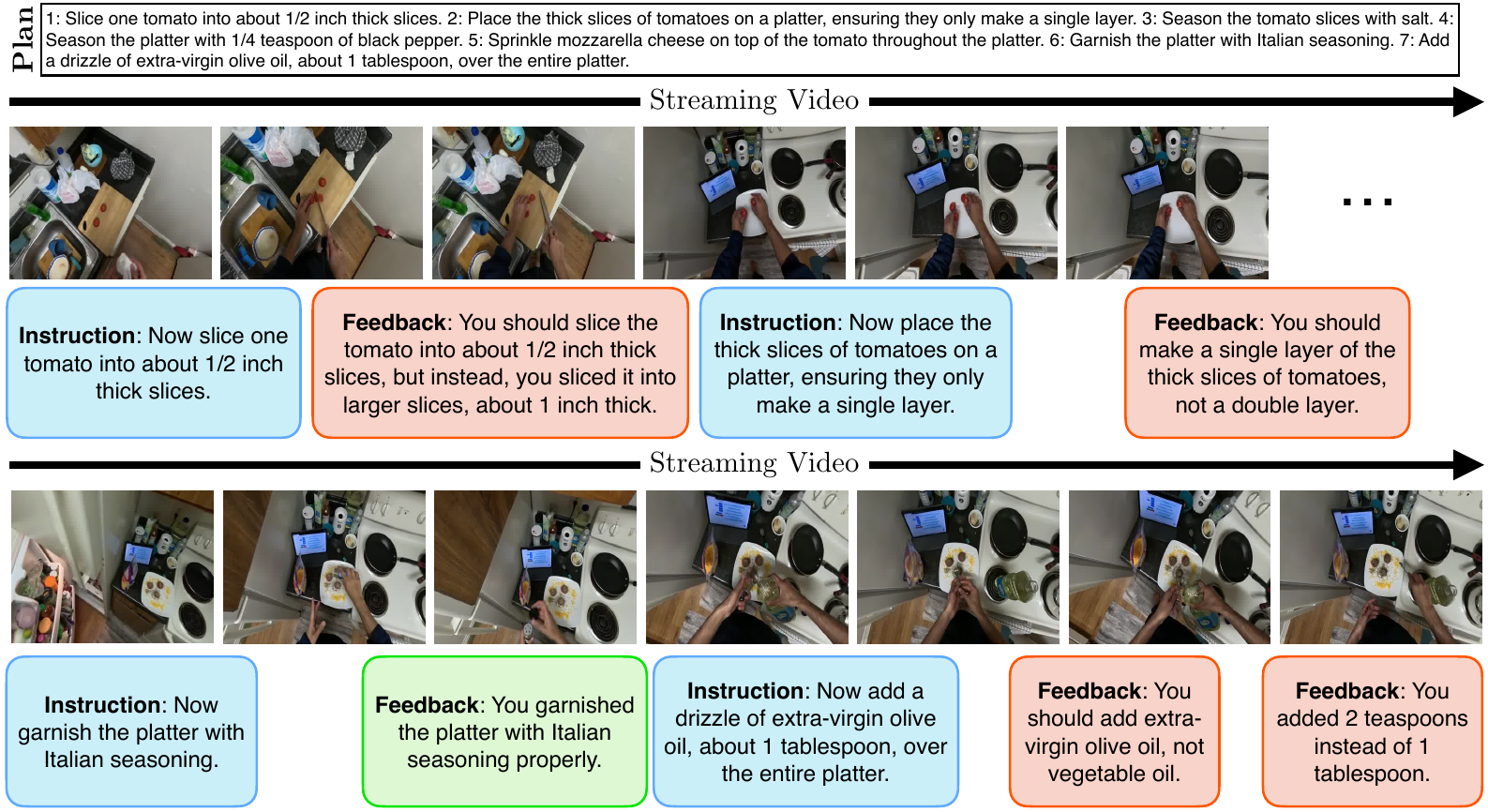}
    \vspace{-0.25cm}
    \caption{An overview of the step-by-step task guidance scenario in our \benchmark{} benchmark and dataset, where the multi-modal LLM provides instructions and feedback that are sufficient to guide the user towards the goal, \eg, making a tomato mozzarella salad (above).}
    \label{fig:teaser}
\end{figure}

However, to be useful as an AI assistant, multi-modal LLMs should be able to guide a user through a task -- ``live'' -- by providing interactive step-by-step instructions. This encompasses three key abilities, i) to provide the next instruction, ii) detect if the instruction has been successfully accomplished by the user, iii) if not, then detect any mistakes made by the user and alert the user as soon as possible. Consider an example where a multi-modal LLM guides a user while making, \eg, Bruschetta in \cref{fig:teaser}. At the stage where the tomatoes are being sliced, an instruction with the desired thickness the tomatoes needs to be provided. Then the model needs to detect whether the user has sliced the tomatoes to the desired thickness, and in the case of a mistake, 
the model needs to alert the user as soon as it observes the mistake.
This calls for the multi-modal LLM to be able to react interactively to events in the video streams. However, current state-of-the-art multi-modal LLMs are still largely limited to turn based interactions \cite{hurst2024gpt,liu2024llavanext,team2024gemini,Qwen2VL,damonlpsg2023videollama} -- they only produce responses when prompted by the user -- or to narration tasks \cite{VideoLLM-online,LION-FS}. 

To address the challenge of live, step-by-step coaching, we introduce the \benchmark{} dataset and benchmark, as currently available large-scale vision-language datasets and benchmarks \cite{Damen2022RESCALING,GraumanWBCFGH0L22,coin-dataset} inadequately capture such interactive scenarios. These existing datasets and benchmarks largely consist of participants recording their daily activities or expert demonstrations, which are insufficient for effectively assessing multi-modal LLMs' ability to provide step-by-step instructions, as they lack the critical scenarios where users make mistakes or diverges from the plan. Therefore, we construct our \benchmark{} benchmark and dataset leveraging the videos from the CaptainCook4D dataset \cite{PeddiACPVGZWKRR24}, specifically because they contain these vital instances of user errors. Each video within \benchmark{} is accompanied by a detailed step-by-step plan along with instructions and feedback, timed appropriately according to the recipe and any mistakes made by the user.

Our contributions in detail are,
\begin{enumerate*}
    \item We introduce the \benchmark{} dataset and benchmark\footnote{\href{https://apratimbh.github.io/livecook/}{Project Page}} by extending the CaptainCook4D dataset \cite{PeddiACPVGZWKRR24}, with timed instruction and feedback messages. The instruction messages describe the next recipe step to follow and the feedback messages are provided to acknowledge successful instruction completions or mistakes made by the user. They are designed to be sufficient to independently guide the user to complete the given recipe.

    \item We introduce the \model{} model, an light-weight multi-modal LLM designed to provide interactive instructions and feedback in cooking scenarios.
    
    \item We evaluate state of the art multi-modal LLMs on our \benchmark{} benchmark, highlighting the strong performance of our \model{} model in such interactive scenarios.
    
\end{enumerate*}

\section{Related Work}

\myparagraph{Datasets for Procedural Activities.}
We provide an overview of related datasets and benchmarks in \cref{tab:dataset_comp}.
The Epic-Kitchens~\cite{Damen2022RESCALING} dataset consists of ego-centric videos that document unscripted, daily activities within a kitchen environment. The Ego-Exo4D~\cite{grauman2023ego} dataset incorporates a considerably diverse array of human activities, as performed by subjects exhibiting with a variety of skill levels.
However, the videos in these datasets are not driven by a specific (multi-step) goal (\cref{tab:dataset_comp}, col 3).
Assembly-101~\cite{SenerCSHSWY22} features videos of people with diverse skill levels assembling and disassembling 101 ``take-apart'' toy vehicles.
HowTo100M~\cite{miech2019howto100m} and COIN~\cite{coin-dataset} provides narrated instructional videos, featuring a variety of activities.
The YouCook2~\cite{zhou2018towards} dataset contains cooking videos of diverse recipes.
However, the subjects in these datasets are usually experts and the videos do not contain mistakes, unlike our \benchmark{} benchmark and dataset (\cref{tab:dataset_comp}, col 5). Finally, HoloAssist~\cite{WangKRPCABFTFJP23} and WTAG~\cite{BaoYZSBISZC23} include videos where subjects are provided live feedback by tele-operator, that cover certain mistakes or user queries. But unlike \benchmark{} benchmark and dataset, these feedback are not sufficient to independently guide the subject towards completion of the high-level goal. 

\begin{table}[t!] 
    \centering
    \small
    \caption{Comparison to our \benchmark{} dataset and benchmark (Ours): where, \emph{Mult-step Goal Driven} refers to whether the videos are driven by a specific goal (\eg, cooking a recipe); \emph{Step-by-Step Instructions}: whether the videos contain subjects following a set of step-by-step instructions; and \emph{Timed feedback}: whether the subjects receive timed feedback per step that are sufficient to guide the subjects to the goal.} 
    \label{tab:dataset_comp} 
    \begin{tabularx}{\linewidth}{@{} 
        >{\raggedright\arraybackslash}p{3.0cm}
        >{\centering\arraybackslash}p{2.0cm}
        >{\centering\arraybackslash}p{1.75cm}
        >{\centering\arraybackslash}p{1.75cm}
        >{\centering\arraybackslash}p{1.75cm}
        >{\centering\arraybackslash}p{1.75cm}
    @{}}
        \toprule
                &                 & {Multi-step}    & {Step-by-Step}    & {Timed}  & \\
        Dataset & {Domain} & {Goal Driven}  & {Instructions}    & {Feedback} & {Length (hrs)} \\
        \midrule
        Epic-Kitchens~\cite{Damen2022RESCALING} & Cooking & $\times$ & $\times$ & $\times$ & 100\\
        Ego-4D~\cite{GraumanWBCFGH0L22} & Daily-life & $\times$ & $\times$ & $\times$ & 3670\\
        Ego-Exo4D~\cite{grauman2023ego} & Daily-life & $\times$ & $\times$ & $\times$ & 1422\\
        Assembly-101~\cite{SenerCSHSWY22} & Toy Assembly & \checkmark & $\times$ & $\times$ & 513\\
        HoloAssist~\cite{WangKRPCABFTFJP23} & Obj. manip. & \checkmark & $\times$ & $\times$ & 166\\
        HowTo100M~\cite{miech2019howto100m} & Diverse & \checkmark & \checkmark & $\times$ & 134k\\
        COIN~\cite{TangDRZZZL019} & Diverse & \checkmark & \checkmark & $\times$ & 512\\
        YouCookv2~\cite{zhou2018towards} & Cooking & \checkmark & \checkmark & $\times$ & 176 \\
        WTAG~\cite{BaoYZSBISZC23} & Cooking & \checkmark & \checkmark & $\times$ & 10\\
        QEVD~\cite{PanchalBB0BD0LM24} & Fitness & $\times$ & \checkmark &  \checkmark & 474\\
        \midrule
        Ours & Cooking & \checkmark & \checkmark & \checkmark & 94\\
        \bottomrule
    \end{tabularx}
\end{table}

\myparagraph{Multi-modal Large Language Models.} Vision language models have seen amazing progress in recent years following the language modeling breakthroughs. Earlier efforts on contrastive learning of vision and language representations led to CLIP~\cite{radford2021learning} like architectures~\cite{jia2021scaling, li2021align, li2022blip}. However, more recent works have reformulated vision tasks as text generation tasks giving rise to a diverse array of large multi-modal language models~\cite{achiam2023gpt, bai2023qwen, grattafiori2024llama,  team2023gemini, team2024gemini, Qwen2VL} building on the success of large language models (LLMs). Early works in this space like Flamingo~\cite{alayrac2022flamingo} leveraged pre-trained language models and vision adapters to align the language and visual representational spaces. Follow up works like Llava~\cite{LiuLWL23a}, InstructBLIP~\cite{dai2023instructblip} instruction tuned these models to enable excellent multi-modal dialogue capabilities. These works were primarily applied to image data which have since been generalized to videos~\cite{ ataallah2024minigpt4, jin2024video, li2023videochat, li2024llama,maaz-etal-2024-video,ZhangLB23}. While these models can take videos as input, they only enable turn based interaction, \ie, the models usually answer a question or narrate the whole video lacking the ability to interactively respond to events in the video unlike our \model{} model.

\myparagraph{Streaming Video Large Language Models.} Many recent works have started to explore online video understanding with large multi-modal language models. VideoLLM-online~\cite{VideoLLM-online} proposes a framework to enable online dialogue over videos with LLMs and use it to train models to narrate long streaming videos. ReKV~\cite{di2025streaming} proposes a training free approach to enable existing video language models to solve streaming visual question answering (StreamVQA). Similarly, TimeChat-Online~\cite{yao2025timechat}, StreamChat~\cite{liu2024streamchat}, LiveCC~\cite{chen2025livecc}, Flash Vstream~\cite{zhang2024flash}, StreamMind~\cite{ding2025streammind}, LION-FS~\cite{LION-FS} introduce various innovations to make multi-modal language models to work on streaming video tasks. This has also prompted development of many streaming video benchmarks~\cite{bhattacharyya2023look,li2025ovo, PanchalBB0BD0LM24, pourreza2025can, wang2025omnimmi, yang2025svbench}. However, all of the previous benchmarks deal with different forms of VQA where the model is asked questions paired with streaming video in natural settings whereas we deal with multi-step interactive videos where the model needs responds to the visual input without being prompted in a goal-directed setting.

\begin{table}[t!] 
    \centering
    \small
    \caption{Dataset statistics: where, the \textit{Followed Feedback} and \textit{Divergent Feedback} indicate whether the action associated with the feedback follows the given instruction or not.} 
    \label{tab:dataset_stats} 
    \resizebox{\linewidth}{!}{
    \begin{tabularx}{\linewidth}{
    @{} Xccccccc @{}
    }
        \toprule
              & & & & {Followed} & {Followed} & {Divergent} & {{Divergent}} \\
              & {Total Length} & {Number of} &  {Number of}   & {Success} & {Mistake} & {Success} & {Mistake} \\
        Split & {(hours)}      & {Videos}    & {Instructions} & {Feedback} & {Feedback} & {Feedback} & {Feedback} \\
        \midrule
        \multicolumn{8}{@{}c}{\emph{Main Set}} \\ \midrule
        Training & 52.4 & 213 & 2913 & 2394 & 686 & &  \\
        Validation & 15.7 & 62 & 861 & 659 & 257 & &  \\
        Testing  & 26.4 & 109 & 1489 & 1135 & 445 & &  \\
        \midrule
        \multicolumn{8}{@{}c}{\emph{Advanced Planning Set}} \\ \midrule
        Training & 51.5 & 209 & 2888 & 2119 & 423 & 244 & 229 \\
        Validation & 14.3 & 57 & 781 & 524 & 133 & 78 & 84 \\
        Testing  & 7.8 & 36 & 481 & 123 & 144 & 115 & 119 \\
        \bottomrule
    \end{tabularx}}
\end{table}

\section{\benchmark{} Benchmark and Dataset}
Here we introduce the \benchmark{} benchmark and dataset to evaluate the ability of multi-modal LLMs to provide step-by-step instructions, focusing on the cooking domain. 
The \benchmark{} benchmark and dataset uses the videos 
from the CaptainCook4D \cite{PeddiACPVGZWKRR24} dataset as it contains actions with mistakes.
This allows us to create a setup akin to a ``non-reactive simulation'', where we task the multi-modal LLM to produce the right instruction and feedback at the appropriate time, but the subject is non-compliant.
Such a setup still provides us with useful insight into ability of multi-modal LLMs to provide step-by-step task guidance, as fully reactive setups are not possible with offline datasets.
We provide statistics of the \benchmark{} benchmark and dataset in \cref{tab:dataset_stats}, including the total length in hours, number of videos, and numbers of instructions and feedback messages.

\myparagraph{Instruction and Feedback Protocol.} The instructions and feedback are annotated using the following protocol, starting from a step-by-step plan. 
The first instruction of the plan occurs at the beginning of the video. If the user makes a mistake while completing the instruction, our benchmark and dataset contains a corresponding feedback just after it occurs. If an instruction is completed successfully, it is acknowledged.
As long as the subject tries to complete the given instruction, irrespective of mistakes, we provide the instruction for the next step.
If the user performs actions which are not aligned with the instruction, \eg, performing recipe steps out of order, we update the step-by-step plan, \eg, by repeating the instruction (see advanced planning set).
Once all the steps are completed, we acknowledge the completion for the whole plan.

\myparagraph{Annotation Process.}
To generate instructions and feedback we leverage the annotations for temporal action segments with action descriptions in the CaptainCook4D dataset. We also leverage the mistake descriptions for the actions containing mistakes. First, we begin by removing noisy mistake annotations and then we annotate the timestamps where those mistakes occur. This allows us to annotate timely feedback for each mistake. We do this for all types of mistakes except {\tt order error} and {\tt missing steps}. Those mistakes reflect the cases where the subject ignores the provided instruction, \ie, they perform a recipe step out of order or ignore the recipe step completely.
Such cases require more complex reasoning from the multi-modal LLM, as the model needs to reason using the information of the instructions completed in the past and the remaining future steps.
Thus, we propose two sets of our \benchmark{} benchmark and dataset: Main Set and Advanced Planning Set. The training, validation, and testing splits within each set follow the original video recoding split from CaptainCook4D, and the test splits correspond to the \benchmark{} benchmark. 

\emph{Main Set.} This set reflect the case where the user largely follows the given instruction. To build the step-by-step plan for this set we leverage the order of the actions in the video.
First we sort the actions by their start time. Normally one action forms one step in the plan. In cases where actions are performed in parallel (compound actions), we group those actions into the same step. 
In this set, instructions are to be given according to their order in the plan, thus eliminating the need for complex reasoning to provide the next instruction.

\emph{Advanced Planning Set.} This set targets scenarios where the user diverges from the prescribed sequence, such as performing steps out of order. To construct it, we use the graph-structured recipe for each CaptainCook4D video to determine the correct step order. When the user does not follow the given instruction, we notify them and identify the action being performed based on the initial plan. The future instruction sequence is then updated to reflect this action (examples in the supplementary material). The test split includes only videos where the user diverges from the initial plan at least once. To reduce complexity, we further restrict the test set to videos without compound actions.

\begin{figure}[t]
    \centering
    \includegraphics[width=\textwidth]{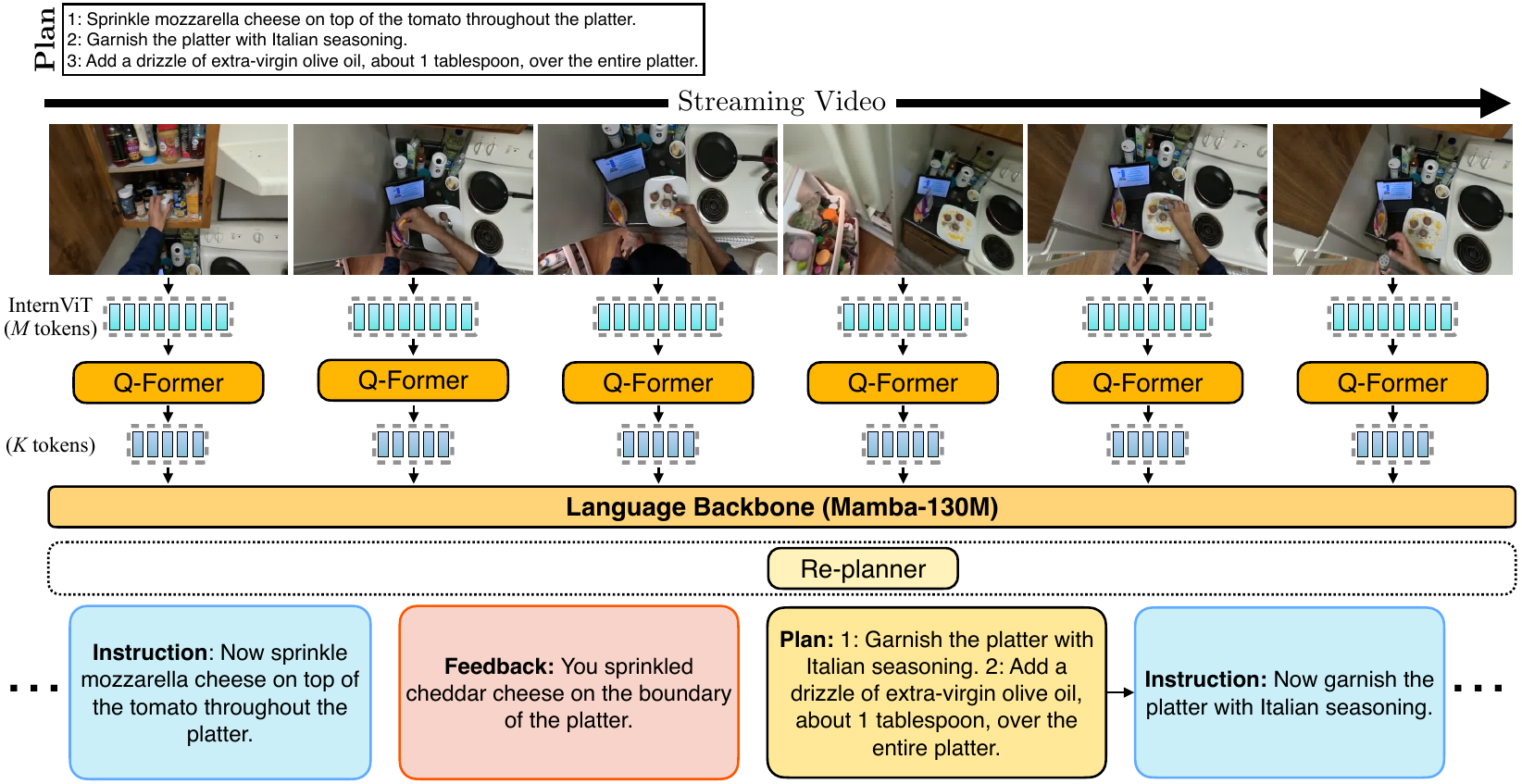}
    \vspace{-0.5cm}
    \caption{Our \model{} model architecture. The input video stream is processed by an InternViT vision head which produces $M$ tokens, and is then reduced to $K$ tokens by a Q-Former. The language backbone produces feedback and invokes the Re-planner if necessary before the next instruction.}
    \label{fig:arch}
\end{figure}

\section{\model{} for Step-by-Step Instructions}
To deal with the challenge of guiding a user through step-by-step instructions we propose our \model{} model. It is a vision-language model with a vision encoder and language model backbone to produce step-by-step instructions and feedback at the appropriate time-step. Given a plan with a set of step-by-step instructions, the \model{} model, shown in \cref{fig:arch}, provides the next instruction to the user, waits for the user to complete the instruction or make a mistake and provide a feedback message. To enable this interactive mode of operation, our \model{} has the following key novel features:
\begin{enumerate*}
    \item It knows ``when to speak'', \ie, it is trained to respond at the appropriate time step when the user has successfully completed the provided instruction or has made a mistake. 
    \item To recognize the diverse set of possible mistakes that could be made by the user, the \model{} model is trained using a novel data augmentation scheme during the fine-tuning phase.
    \item To provide the next relevant instruction, the \model{} model re-plans using an external re-planning module, in case the user diverges from the provided instruction and performs a step out of order.
\end{enumerate*}
Furthermore, the \model{} incorporates a recurrent Mamba-130M~\cite{abs-2312-00752} language model backbone to enable efficient training and inference over long video streams. Next, we discuss the \model{} architecture and training details.

\subsection{\model{} Architecture}
The \model{} architecture consists of a vision encoder and a light-weight but powerful language backbone. 
A key consideration of the architecture of our \model{} model is efficiency. 
For real-world application of our \model{} model, inference using our model should ideally be possible using compute constrained edge devices such as mobile phones or smart glasses.
Reliance on external servers with larger compute capabilities is both costly and comes with additional latency issues.
A second key consideration is that the cooking activities in the \benchmark{} benchmark requires recognition of fine-grained actions and small objects, \eg, adding the correct amount of salt, coating a cup vs a bowl with oil.
Next, we describe these components in more detail.

\myparagraph{Vision Encoder.} The vision encoder consists of a InternViT-300M-448px-V2\_5 \cite{chen2024internvl} head along with a Q-Former \cite{0008LSH23} based adapter. As shown in \cref{fig:arch} the InternViT head produces $M$ tokens per input video frame.
As we use a light-weight language backbone, we do not want to offload raw visual processing to the language backbone and thus employ a powerful Q-Former with multiple transformer layers. 
For every input video frame, the Q-Former reduces the $M$ tokens to $K$ tokens which are then input to the language backbone.
To generate these $K$ tokens, the Q-Former uses four cross attention layers to gradually blend in the important visual information from the input video frame. These tokens are then used as input to the language backbone, interleaved with the text tokens corresponding to the step-by-step plan, instructions and feedback.

\myparagraph{``When-to-Say''.} To respond at the appropriate timestep, following \cite{VideoLLM-online,PanchalBB0BD0LM24} our \model{} model uses two special tokens <vision> and <response>. The <vision> token allows the \model{} model to interactively ask for the next video frame as input, while <response> allows the \model{} to respond with an instruction or a feedback at the appropriate time.

\myparagraph{Iterative Re-planning.} To determine the next instruction, we employ the following strategy. 
If the user executes the current instruction—whether correctly or with errors—the subsequent step from the original plan is issued. 
On the other hand, if the user diverges from the prescribed sequence and performs an out-of-order step, our \model{} invokes an external re-planner (see \cref{fig:arch}). 
The re-planner receives as input the initial step-by-step plan, the set of completed steps, and feedback generated by \model{}, and then selects the next optimal step to ensure successful task completion. 
In yet another scenario, if the intended step is skipped, the re-planner evaluates whether repeating the current instruction is appropriate given the user’s progress. 
Similarly, if a step is completed prematurely, the re-planner determines whether to remove it from the remaining plan.
This iterative and selective re-planning mechanism provides two principal advantages. 
First, it enables selective use of large-scale models with enhanced reasoning capabilities only when necessary; in our implementation, Qwen3-32B~\cite{abs-2505-09388} serves as the re-planner. 
Second, it eliminates the need for \model{} to internally store and recall the entire plan, allowing it to focus exclusively on delivering accurate and timely feedback (more details in the appendix).

\subsection{\model{} Training}\label{sec:training_stages}
We now provide details of our training scheme, which is composed of two stages.

\myparagraph{Pre-training.} This stage aligns the Q-Former's vision embeddings with the language backbone's text embeddings by training only the Q-Former adapter. We use a diverse set of image and video datasets to instill two key visual skills:
First, for object grounding, \model{} is trained on image datasets like LVIS \cite{GuptaDG19} (for diverse household objects) and on video data using VISOR annotations from EPIC-KITCHENS \cite{Damen2022RESCALING}. Tasks include image/video captioning, object recognition, and bounding box prediction.
Second, for fine-grained action understanding, \model{} learns action recognition from SSv2 \cite{Goyal_2017_ICCV} and video narration from EPIC-KITCHENS and Ego4D. Narration helps ground the model to actions in these large-scale egocentric datasets, particularly cooking activities.

\myparagraph{Fine-tuning.} During the fine-tuning phase the \model{} model is trained to provide the appropriate instructions and feedback at the appropriate time, using the <vision> and <feedback> special tokens as described above. 
At this stage, both the Q-Former's based adapter and the language backbone is trained. 
The \model{} needs to recognize the successful completion of instructions and mistakes. To this end, in addition to training on the step-by-step instructions and feedbacks from our \benchmark{} dataset, we apply several augmentations as described below.

\subsection{\model{} Augmentation}
We now provide details of our data augmentation scheme during the fine-tuning phase.

\myparagraph{Temporal Augmentation.} To maintain temporal accuracy of feedbacks while dealing with long videos in the \benchmark{} benchmark, we introduce temporal jittering during training. Specifically, we jitter the starting timestamp of each instruction by a constant $\pm$K seconds. This temporal jittering deals that fact that predictions by autoregressive models, \eg, our \model{}, can accumulate errors. Jittering the starting timestamp of an instruction ensures that the \model{} can successfully predict feedbacks irrespective of the previous accumulated error. In practice, we find $K=30$ to work well.

\myparagraph{Instruction Completion Augmentation.} To help the \model{} recognize the successful completion of instructions, we augment our training set by converting videos from the EPIC-KITCHENS and Ego4D datasets to the step-by-step instruction and feedback format of our \benchmark{} dataset. In detail, we consider the action descriptions in the EPIC-KITCHENS dataset and the Ego4D Goal-Step datasets and use a Qwen2.5-8B model \cite{abs-2412-15115} to convert these action descriptions to instructions. We provide the instruction at the action start time and feedback message at the action end timestamp. As these datasets do not contain any mistakes, the feedback messages acknowledge the successful completion of the instruction.

\myparagraph{Counterfactual Mistake Augmentation.} Recognizing mistakes and providing timely feedback is a key challenge of our \benchmark{} benchmark. To this end, we formulate a novel data augmentation scheme to generate (counterfactual) mistakes in the EPIC-KITCHENS and Ego4D datasets. First, we convert the action description to plausible grounded counterfactual action descriptions. These grounded counterfactual action descriptions are used to generate instructions and thus construct scenarios where the user tries to follow the given instruction but makes a mistake (more details in the appendix). To help recognize divergent mistakes, where the user does not follow the provided instruction, we augment the fine-tuning dataset by swapping instructions between recipe steps. The feeback is constructed to explicitly state that the user did not follow the provided instruction and instead performed a different action. This format of feedback triggers the re-planning module during the inference stage (more details in the appendix).

\begin{table}[t!]
    \centering
    \small
    \caption{Zero-shot evaluation on the main set of the \benchmark{} benchmark.} 
    \label{tab:zeroshot} 

    \begin{tabularx}{\linewidth}{@{}
        >{\raggedright\arraybackslash}p{3.5cm} 
        >{\centering\arraybackslash}p{2.0cm}   
        @{\hspace{0.75cm}}
        >{\centering\arraybackslash}p{1.0cm}   
        >{\centering\arraybackslash}p{1.0cm}   
        >{\centering\arraybackslash}p{1.0cm}   
        >{\centering\arraybackslash}p{1.0cm}   
        >{\centering\arraybackslash}p{1.6cm}   
        @{}}
        \toprule
         & Instruction & \multicolumn{5}{c}{Mistake} \\
         \cmidrule{2-2} 
         \cmidrule{3-7}
        Method & {IC-Acc$\uparrow$} & Prec.$\uparrow$ & Rec.$\uparrow$ & F1$\uparrow$ & BERT$\uparrow$ & ROUGE-L$\uparrow$ \\
        \midrule
        LLaVA-NeXT~\cite{liu2024llavanext}           & 1.4 & 0.00 & 0.00 & 0.00 & 0.000 & 0.000\\
        Video-ChatGPT~\cite{maaz-etal-2024-video}     & 1.6 & 0.00 & 0.00 & 0.00 & 0.000 & 0.000\\
        VideoChat2~\cite{li2024mvbench}               & 1.6 & 0.00 & 0.00 & 0.00 & 0.000 & 0.000\\
        Video-LLaVA~\cite{zhu2023languagebind, lin2023video} & 2.0 & 0.00 & 0.00 & 0.00 & 0.000 & 0.000\\
        VideoLLaMA3-7B~\cite{damonlpsg2025videollama3}  & 1.8 & 0.00 & 0.00 & 0.00 & 0.000 & 0.000\\
        Videollm-online~\cite{VideoLLM-online} & 0.03 & 0.02 & \textbf{0.98} & 0.04 & 0.332 & 0.248 \\
        Qwen2-VL-7B~\cite{Qwen2VL} & 6.3 & 0.02 & 0.69 & \textbf{0.05} & {0.377} & {0.256}\\
        
        Qwen2.5-VL-7B~\cite{abs-2502-13923} & {18.9} & \textbf{0.18} & 0.01 & 0.02 & {0.299} & {0.219}\\
        Gemini-2.5-Flash \cite{abs-2507-06261} & \textbf{23.1} & 0.01 & 0.22 & 0.02 & \emph{0.410} & \emph{0.342}\\
        \bottomrule
    \end{tabularx}
\end{table}

\section{Experiments}
We begin with an introduction to the metrics used to evaluate models for interactive step-by-step task guidance followed by the experimental results.

\subsection{Evaluation Metrics.} We use the following metrics to measure both the ability of the models to detect successfully completed instructions and to provide feedback when mistakes occur.

\myparagraph{Instruction Completion Accuracy (IC-Acc).} IC-Acc measures the proportion of instructions that the model correctly detects as successfully completed by the user. Specifically, this requires that the model provides the correct instruction, the user completes it, and the model identifies this completion. To mitigate temporal annotation noise, we consider a prediction correct if it falls within a small window centered on the ground-truth completion time. In practice, a 30-second window is sufficient: it typically spans the last $\sim$25\% of the current step and the first $\sim$25\% of the next step in the \benchmark{} benchmark, balancing robustness to noise with accuracy.

\myparagraph{Mistake Detection Precision (Prec.), Recall (Rec.) and F1.} 
To calculate mistake detection Precision, Recall, and F1 scores in our interactive streaming setup, we define:
\begin{itemize*}[nosep, leftmargin=*]
    \item \emph{Mistake True Positive}: A mistake detected by the model within a small temporal window centered on the timestamp of a ground truth mistake.
    \item \emph{Mistake False Negative}: A ground truth mistake that the model fails to detect within this temporal window.
    \item \emph{Mistake False Positive}: A mistake detected by the model when no corresponding ground truth mistake occurs within the temporal window.
    \item \emph{Mistake True Negative}: An instruction correctly followed by the user (no ground truth mistake) where the model correctly detects no mistake.
\end{itemize*}
We use the same temporal window size as in the IC-Acc metric.

\myparagraph{Mistake Feedback Fluency (BERT and ROUGE-L).} To measure the fluency of the models in providing appropriate feedback, we use the ROUGE-L and BERT scores. We only consider the fluency of feedback provided in case true positive mistake detections. That is, when the feedback is provided within the temporal window of a ground truth feedback as described above. Importantly, these scores are only meaningful when comparing models with similar true positive detection rates, since differences in detection accuracy can confound fluency comparisons. This is because models with lower detection rates produce fewer feedback instances, which can skew the distribution of ROUGE and BERT scores and make fluency appear artificially higher or lower.

\begin{table}[t!]
    \centering
    \caption{Evaluation of fine-tuned models on the \benchmark{} benchmark (\,$^{\dagger}$indicates models fine-tuned by us).} 
    \label{tab:finetune} 
    \small
    \begin{tabularx}{\linewidth}{@{}
        >{\raggedright\arraybackslash}p{3.5cm} 
        >{\centering\arraybackslash}p{2.0cm}   
        @{\hspace{0.75cm}}
        >{\centering\arraybackslash}p{1.0cm}   
        >{\centering\arraybackslash}p{1.0cm}   
        >{\centering\arraybackslash}p{1.0cm}   
        >{\centering\arraybackslash}p{1.0cm}   
        >{\centering\arraybackslash}p{1.6cm}   
        @{}}
        \toprule
         & Instruction & \multicolumn{5}{c}{Mistake} \\
         \cmidrule{2-2} 
         \cmidrule{3-7}
        Method & {IC-Acc$\uparrow$} & Prec.$\uparrow$ & Rec.$\uparrow$ & F1$\uparrow$ & BERT$\uparrow$ & ROUGE-L$\uparrow$ \\
        \toprule
        \multicolumn{7}{c}{\emph{Main Set}} \\
        \midrule
        Videollm-online$^\dagger$ \cite{VideoLLM-online} & 7.6 & 0.04 & 0.01 & 0.01 & 0.434 & 0.412 \\
        \model{} (w/o-ICAug) & 7.8 & 0.05 & 0.01 & 0.01 & 0.605 & 0.542 \\
        \model{} (w/o-CFAug) & 14.3 & 0.12 & 0.03 & 0.05 & 0.558 & 0.511\\
        \model{} (Ours) & \textbf{31.5} & \textbf{0.17} & \textbf{0.10} & \textbf{0.13} & \emph{0.651} & \emph{0.561}\\
        \midrule
        \multicolumn{7}{c}{\emph{Advanced Planning Set}} \\
        \midrule
        \model{} (w/o-reP) & 10.9 & \textbf{0.38} & 0.10 & 0.16 & 0.912 & 0.901 \\
        \model{} (Ours) & \textbf{12.6} & \textbf{0.38} & \textbf{0.13} & \textbf{0.19} & \emph{0.941} & \emph{0.927}\\
        \bottomrule
    \end{tabularx}
\end{table}

\begin{table}[t!]
    \centering
    \caption{Turn-based evaluation of on the main set of the \benchmark{} benchmark.} 
    \label{tab:turnbased} 
    \small
    \begin{tabularx}{\linewidth}{@{}
        >{\raggedright\arraybackslash}p{3.5cm} 
        >{\centering\arraybackslash}p{2.0cm}   
        @{\hspace{0.75cm}}
        >{\centering\arraybackslash}p{1.0cm}   
        >{\centering\arraybackslash}p{1.0cm}   
        >{\centering\arraybackslash}p{1.0cm}   
        >{\centering\arraybackslash}p{1.0cm}   
        >{\centering\arraybackslash}p{1.6cm}   
        @{}}
        \toprule
         & Instruction & \multicolumn{5}{c}{Mistake} \\
         \cmidrule{2-2} 
         \cmidrule{3-7}
        Method & {IC-Acc$\uparrow$} & Prec.$\uparrow$ & Rec.$\uparrow$ & F1$\uparrow$ & BERT$\uparrow$ & ROUGE-L$\uparrow$ \\
        \toprule
        
        VideoLLaMA3-7B~\cite{damonlpsg2025videollama3}  & 17.8 & 0.08 & \textbf{0.61} & \textbf{0.15} & 0.406 & 0.346\\
        Qwen2-VL-7B~\cite{Qwen2VL} & 19.4 &  0.06 & 0.46 & {0.11} & \emph{0.398} & \emph{0.293}\\
        Qwen2.5-VL-7B~\cite{abs-2502-13923} & \textbf{38.9} & \textbf{0.11} & 0.04 & 0.06 & {0.348} & 0.230\\
        \midrule
        \model{}$^{\dagger}$ (Ours) & \textbf{51.0} & \textbf{0.22} & {0.17} & \textbf{0.19} & \emph{0.631} & \emph{0.535} \\
        \bottomrule
    \end{tabularx}
\end{table}

\subsection{Zero-Shot Evaluation}
State of the art multi-modal LLMs are usually limited to turn-based interactions. 
Applying such models to the streaming setup of the \benchmark{} benchmark is thus highly challenging.
Therefore, we need to employ a special online prompting strategy. 
This strategy involves prompting the model at regular intervals to detect both successful completions of instructions and mistakes. However prompting after every input frame is not computationally feasible. To balance accuracy and compute requirements, we prompt the models at an interval of 5 seconds. 
We provide details of the prompts in the appendix. 

On the other hand, streaming multi-modal LLMs such as Videollm-online \cite{VideoLLM-online} are targeted at providing online narrations of input video streams. Such models are therefore not directly applicable to our \benchmark{} benchmark. Thus, to evaluate such models, we first ask the model to generate online narrations for the whole video. We then feed these narrations in an online manner to a ``helper'' LLM that given the narrations and the action instruction, predicts if the action is completed or not. If ``yes'', we move on to the next instruction. If ``not'', we ask the ``helper'' LLM to predict if the narrations suggest a mistake has been made and provide feedback for correction. We use Phi-3-mini-4k-Instruct~\cite{abs-2404-14219} as the ``helper'' LLM.

We report the zero-shot evaluation results in \cref{tab:zeroshot} for the main set of the \benchmark{} benchmark. Overall, Gemini-2.5-Flash \cite{team2024gemini} performs best. It can recognize 18.9\% of instructions being successfully completed by the user. Models such as VideoLLaMA3-7B~\cite{damonlpsg2025videollama3}, Video-LLaVA~\cite{lin2023video,zhu2023languagebind}, VideoChat2~\cite{li2024mvbench}, Video-ChatGPT~\cite{maaz-etal-2024-video}, LLaVA-NeXT~\cite{liu2024llavanext} have trouble following instructions. Even when prompted to detect if the person has completed a given instruction these models tend to answer ``yes'' too early. This highlights a gap in understanding scenes as they unfold in a streaming setup. This is likely an artifact of the question-answer style training scheme of these models where the entire video is always available to the model. Furthermore, Videollm-online~\cite{VideoLLM-online} predicts narrations which are not fully informative of the action and therefore a mistake is detected very often leading to a very high mistake recall and a very low instruction completion detection accuracy.

Overall, when it comes to mistake detection, none of the zero-shot approaches performs well. Qwen2-VL-7B-Instruct \cite{Qwen2VL} overestimates the occurrence of mistakes, leading to higher recall but low precision. Qwen2.5-VL-7B-Instruct \cite{abs-2502-13923} on the other hand is more precise in detecting mistakes, but with low overall F1 score. This weak performance can be attributed to the fact that detecting mistakes in the \benchmark{} benchmark is very challenging due to a variety of reasons.
First is the fine-grained nature of many mistakes, \ie, taking 1 teaspoon vs 1 tablespoon or 2 teaspoons of sugar, spilling flour on the kitchen counter etc, requires both fine-grained object recognition and action-recognition abilities. This is made additionally challenging by the fact that the model needs to look out of a diverse set of possible mistakes.
Secondly, mistakes need to be detected as soon as they occur (similar to the instruction completions). 
This again exposes the limitations of current models in understanding scenes as they unfold in a streaming setup.

\subsection{Evaluation of Fine-tuned Models}
Here we evaluate our \model{} model fine-tuned on the \benchmark{} dataset. Unlike the zero-shot baselines such as Qwen2.5-VL-7B-Instruct \cite{abs-2502-13923} in \cref{tab:zeroshot}, our \model{} model is fully interactive, \ie, it can decide ``when to say'' to point out successful completion of instructions and mistakes after every input frame. Thus, it does not require expensive prompting strategies. 
We also consider two ablations of our \model{} model on the main benchmark set: 
\begin{enumerate*}
    \item without instruction completion augmentation (w/o-ICAug),
    \item without counterfactual augmentation (w/o-CFAug),
\end{enumerate*}
In case of the advanced planning set, we consider an ablation without an external re-planning module (w/o-rP).

\begin{figure}[t!]
    \centering
    \includegraphics[width=\textwidth]{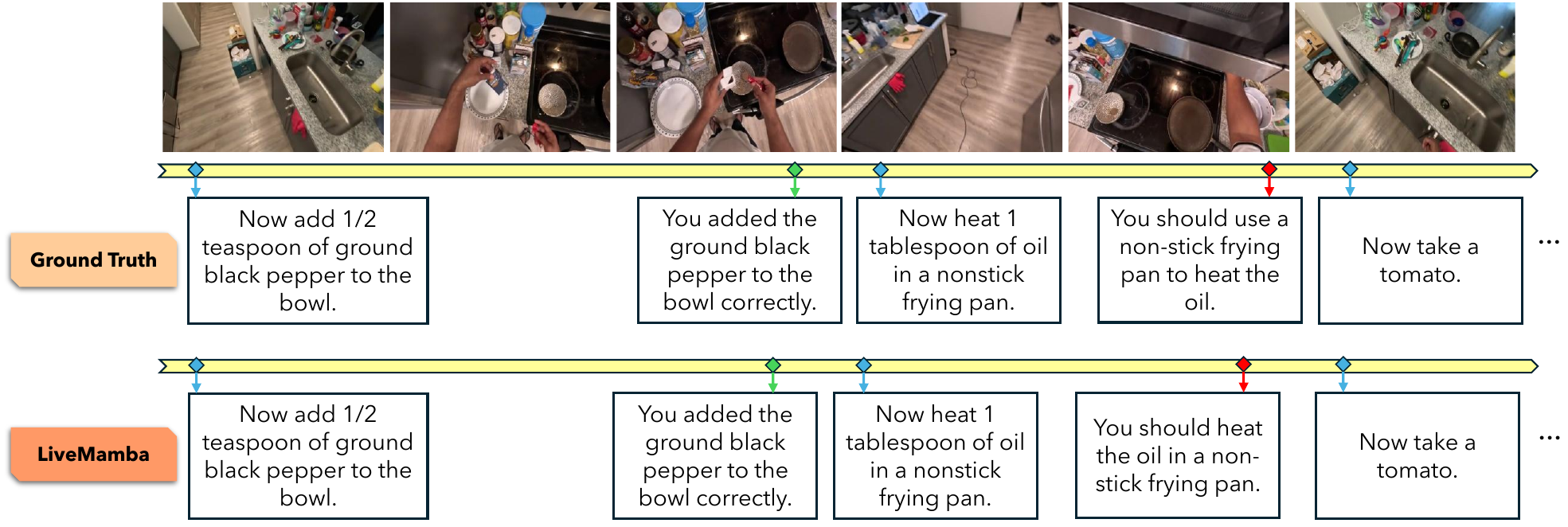}
    \vspace{-0.5cm}
    \caption{Our \model{} is able to successfully recognize the person has added the black pepper as instructed and points out when the person should heat the oil in a non-stick frying pan, in the \benchmark{} benchmark.}
    \label{fig:livemamba_qual_ex1}
\end{figure}

Firstly, in \cref{tab:finetune} we see that our \model{} model significantly improves performance over the zero-shot models in \cref{tab:zeroshot}. We show example predictions of our model in \cref{fig:livemamba_qual_ex1}. 
The weaker performance of the fine-tuned Videollm-online model shows that pre-training only on narration data is not sufficient. 
Furthermore, the use of our efficient Mamba-130M backbone \cite{abs-2312-00752} allows us to use a higher number of embedding tokens per frame (32 vs 10) at similar memory costs compared to transformer based models such as Videollm-online. 
In addition to the pre-training data, the key reasons for the improved performance of the \model{} model is our data augmentation scheme during the fine-tuning phase.
The addition of instruction completion augmentation (ICAug) leads to a significant improvement in IC-Acc from 7.8\% to 14.3\% in the main set along with an improvement in mistake F1 scores. The improvement in mistake F1 score is particularly interesting, as correctly recognizing successful completion of instructions helps the \model{} more precisely point out mistakes.   
Furthermore, the addition of counterfactual mistake augmentation (CFAug) significantly boosts the mistake F1 score from 0.05 to 0.10 in the main set, along with a significant jump in feedback fluency. 
The significant jump in mistake detection accuracy shows that the curation of high quality mistake data is a promising direction for future research.
Furthermore, we see that the use of the external re-planning module leads to better performance on the advanced planning set. The external re-planning module helps the \model{} select the next instruction in case of divergent mistakes over the complex graph structured recipes in the \benchmark{} benchmark from CaptainCook4D. Also note that, our counterfactual (divergent) mistake augmentation scheme significantly boosts the mistake detection scores in this set.
Overall, providing the correct instruction in the advanced planning set remains highly challenging, as shown by the lower IC-Acc scores. Note that, the majority of the mistakes in the advanced planning set are divergent: they point out the intended action given the instruction and the divergent action. Furthermore, the mistake detection scores are calculated only when the provided instruction matches with the groundtruth instruction. Thus, the mistake detection scores are not comparable across the main and advanced planning sets. 

In terms of throughput, on a consumer Nvidia H100 GPU, our LiveMamba model has a real-time factor of 4 on average: it can process input data four times as fast at 8.1 frames per second as it becomes available at 2 frames per second. In terms of latency, that is time to generate the first token, is 1.1 seconds on average. Re-planning using the Qwen3-32B\cite{abs-2505-09388} model takes 6.1 seconds on average. 

\subsection{Turn-based Evaluation}
The streaming evaluation in \cref{tab:zeroshot,tab:finetune} tested models on multi-step user guidance, a setup inherently challenging due to error propagation; for instance, failing to detect instruction completion renders subsequent predictions incorrect. While this reflects real-world scenarios where accurate completion detection is vital for task guidance, such difficulty can make it more challenging to measure progress on the \benchmark{} benchmark. Therefore, we also propose a turn-based evaluation, where models are assessed on each recipe step independently. 

We report the results in \cref{tab:turnbased} on the main set of the \benchmark{} benchmark. Overall we observe much higher IC-Acc scores compared to the streaming setup of \cref{tab:zeroshot,tab:finetune}. Most significantly, we observe higher mistake precision, recall, F1 scores of our fine-tuned \model{} model compared to the zero-shot models. This again shows that fine-tuning on our \benchmark{} dataset along with our data augmentation scheme during fine-tuning allows our \model{} model to better recognize mistakes compared to zero-shot baselines. 
Thus, the turn-based evaluation in \cref{tab:turnbased} along with the streaming setup in \cref{tab:zeroshot,tab:finetune} together provide a more detailed picture of progress on the \benchmark{} benchmark.

\section{Conclusion}
We address the challenge of enabling multi-modal LLMs to provide live, interactive step-by-step guidance. To this end, we introduce \benchmark{} dataset and benchmark, featuring densely annotated, timed instructions and feedback, including for mistakes, timestamped to visual occurrences. We also propose \model{}, a novel streaming multi-modal LLM designed for this task. \model{} utilizes a lightweight Mamba backbone, a "when-to-say" mechanism, novel data augmentation for mistake recognition, and iterative re-planning for adaptive delivery. Evaluations show existing multi-modal LLMs struggle with live task guidance, whereas \model{}, establishes a strong baseline, significantly outperforming others in diverse metrics.

\myparagraph{Limitations.} Our work is focused on the cooking domain through \benchmark{}. While \model{} establishes a strong baseline, detecting subtle mistakes and robustly handling complex scenarios, particularly those involving order errors or missed steps in the advanced planning set, remains challenging, for all state of the art open-source models. 

\myparagraph{Broader Impacts.} Language models can produce harmful and biased content, make incorrect claims and produce wrongful advice. This needs to be taken into account when interacting with, deploying or building on these models, particularly in domains where incorrect advice may lead to physical harm. It also has to be taken into account that any computer vision model processing visual information about human subjects could in principle extract information beyond what is required for the use-case, such as biometric information.

\bibliographystyle{plainnat}
\bibliography{neurips_2025}

\newpage
\appendix
\section*{\centering\Large Appendix}
\section{Overview}

In the following we provide details of the annotation process of our \benchmark{} dataset and benchmark, details of the re-planner used in the \model{} model, training details of our \model{} model, additional qualitative results from our \model{} model, and details of the prompts used for zero-shot evaluation.

\section{\benchmark{} Annotation Details}\label{sec:annotation_details}

Our \benchmark{} dataset and benchmark is built upon the CaptainCook4D dataset~\cite{PeddiACPVGZWKRR24}. The CaptainCook4D dataset contains $384$ videos in total. Each video records a person preparing a dish given a recipe, such as breakfast burritos and tomato mozzarella salad, from an egocentric view. Each video is associated with a graph-structured recipe\anumfootnote{Examples can be found at \url{https://captaincook4d.github.io/captain-cook/recipe.html}.}, and annotated with temporal action segments: action descriptions with the corresponding starting time and ending time. If an action contains a mistake, then there is a description of the mistake. However, there is no timestamp annotation for the mistake. The average duration of action segments in CaptainCook4D is about $52.78$ seconds
, which is about $106$ frames if using frame rate as $2$. This also motivates us to use a light-weight Mamba based model to encode more frames given the hardware constraint.

The CaptainCook4D dataset includes $7$ categories of mistakes: {\tt preparation error}, {\tt technique error}, {\tt measurement error}, {\tt temperature error}, {\tt timing error}, {\tt order error}, and {\tt missing steps}. 
We annotate the timestamps of the mistakes when they just happened. 
We remove some noisy mistake annotations in CaptainCook4D, and annotate for all mistake categories except for {\tt order error} and {\tt missing steps}. 
\cref{tab:mistake_stats} shows the numbers of mistake instances per mistake category. Note, we show the mistake statistics here for a complete description of the \benchmark{} dataset and benchmark. This mistake category information is not used in our experiments presented in the paper.

Our \benchmark{} features with a step-by-step plan, and timestamped instructions and feedback for each video recording. In what follows, we describe how we obtain those for the main set and advanced planning set respectively.

\subsection{Main Set}
\begin{figure}[t]
    \centering
    \includegraphics[width=\textwidth]{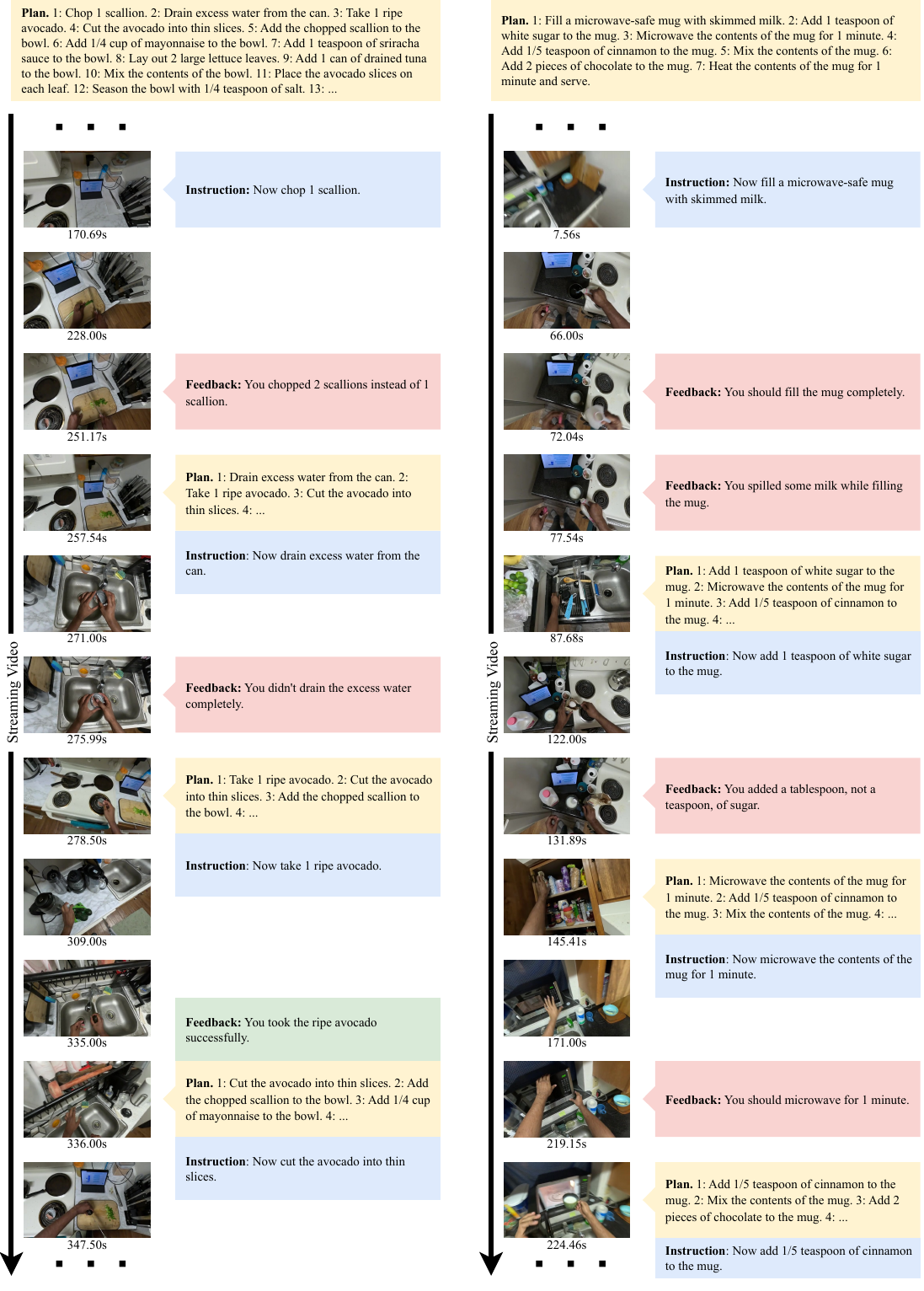}
    \caption{Data samples from the main set. Left: the user prepares spicy tuna avocado wraps. Right: the user prepares spiced hot chocolate.}
    \label{fig:simple_data}
\end{figure}

In the main set (\cref{fig:simple_data}), we assume that the user always tries to follow the given instruction. That is, there is no action order error. Given a video, we first sort the actions by their starting time in ascending order (primary key) and ending time in descending order (secondary key). 
Based on this order, we build the step-by-step plan. Usually one action description forms one step in the plan. In cases where one action is temporally contained in another action (actions performed in parallel), we group those action descriptions into one step. 
Accordingly, we create action groups, where one step in the plan corresponds to one action group.
The first step in the plan provides the contents for the next instruction to be given.

Given a video, we treat the first action's starting time as the video starting time and the first instruction is given at that time.
Once an action finishes, its description is removed from the step-by-step plan. 
Once all the actions in the current action group finish, the current step in the plan will be empty. We remove the empty step in the plan 
and give the instruction for the next step.
The instruction is a sentence containing all the information in the action descriptions in the step.
Once an action finishes successfully, we acknowledge the success via summarizing the description of the just finished action, and using the words like successfully, correctly, properly, and etc. This type of feedback is given at the action's ending time.
If an action contains a mistake, we give the feedback regarding the mistake at our annotated timestamp. The feedback is a sentence pointing out the mistake based on the mistake description given in CaptainCook4D.
Once all the steps in the initial plan are completed, we output the feedback ``{\tt You have finished all the steps.}'' to acknowledge the completion. 
\cref{fig:simple_data} shows data samples from the main set. Whenever an action finishes, it is removed from the step-by-step plan.
The examples show that recognizing mistakes requires fine-grained understanding of user actions and objects in the scene across long time-horizons.

\begin{table}[t] 
    \centering
    \small
    \caption{Mistake category statistics.
    } 
    \label{tab:mistake_stats} 
    \begin{tabularx}{\linewidth}{@{} Xccccc @{}}
        \toprule
        Split & {Preparation Err.} & {Technique Err.} & {Measurement Err.} & {Temperature Err.} & {Timing Err.} \\
        \midrule
        \multicolumn{6}{@{}c}{\emph{Main Set}} \\ \midrule
        Training & 198 & 224 & 159 & 27 & 78 \\
        Validation & 63 & 79 & 68 & 10 & 37 \\
        Testing  & 110 & 156 & 113 & 17 & 49 \\
        \midrule
        \multicolumn{6}{@{}c}{\emph{Advanced Planning Set}} \\ \midrule
        Training & 182 & 216 & 152 & 27 & 75 \\
        Validation & 54 & 66 & 55 & 10 & 32 \\
        Testing  & 68 & 85 & 74 & 10 & 26 \\
        \bottomrule
    \end{tabularx}
\end{table}

\begin{table}[t!] 
    \centering
    \small
    \caption{Re-plan statistics on the advanced planning set (averaged over videos requiring re-planning in each split).} 
    \label{tab:replan_stats} 
    \begin{tabularx}{\linewidth}{@{} X S[table-format=0.0] S[table-format=0.0] S[table-format=0.0] S[table-format=0.0] @{}}
        \toprule
        & {Number of Videos} & {Average Video} & {Average Number of} & {Average Number of} \\
        Split & {Requiring Re-planning} & {Length (minutes)} & {Instructions} & {Re-plan Steps} \\
        \midrule
        Training & 94 & 12.8 & 13.2 & 4.3 \\
        Validation & 29 & 13.2 & 13.6 & 4.8 \\
        Testing  & 36 & 12.9 & 13.4 & 6.1 \\
        \bottomrule
    \end{tabularx}
\end{table}

\subsection{Advanced Planning Set}
\begin{figure}[t]
    \centering
    \includegraphics[width=\textwidth]{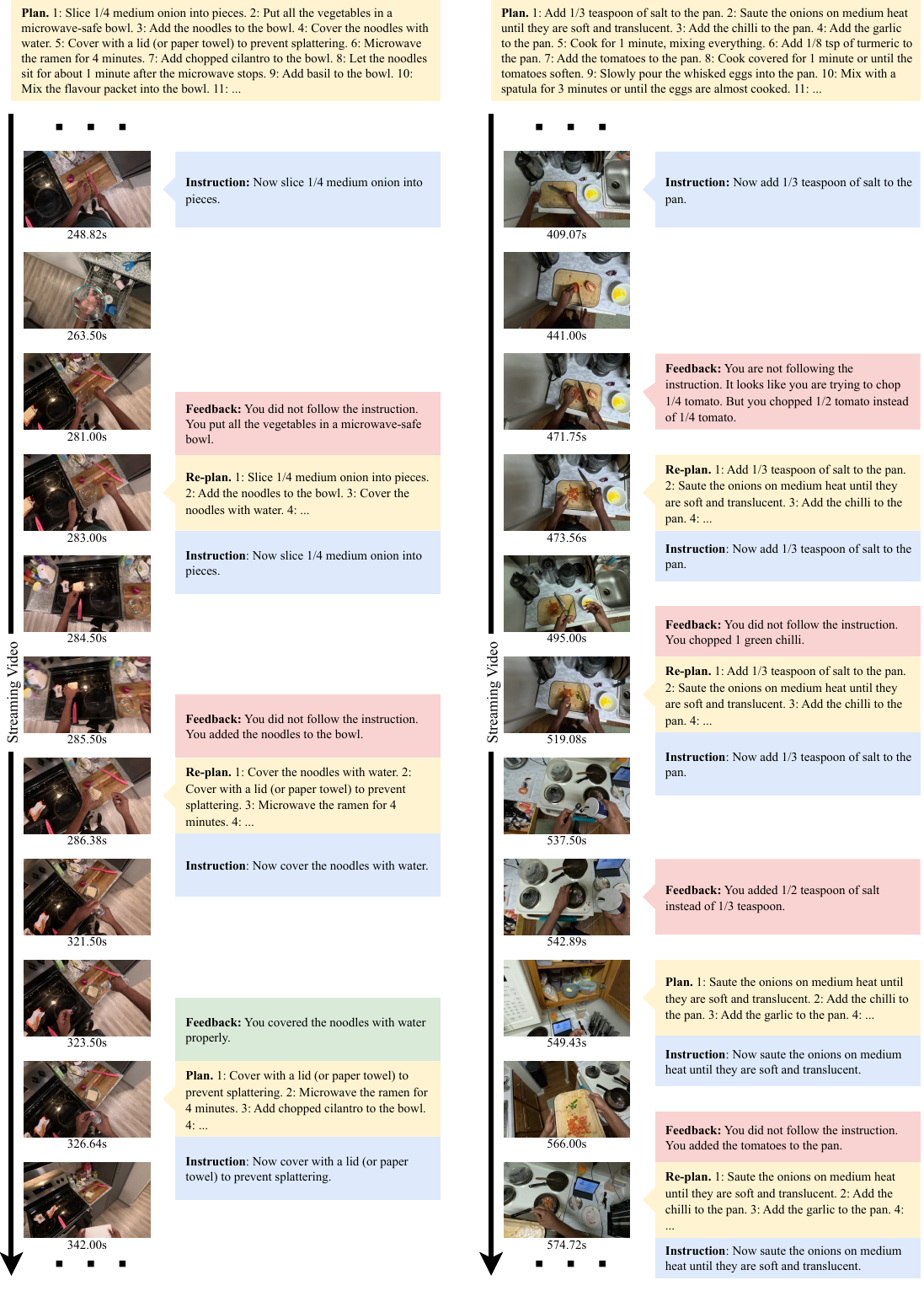}
    \caption{Data samples from the advanced planning set. Left: the user is making ramen. Right: the user is preparing scrambled eggs.}
    \label{fig:advanced_data_1}
\end{figure}

\begin{figure}[t]
    \centering
    \includegraphics[width=\textwidth]{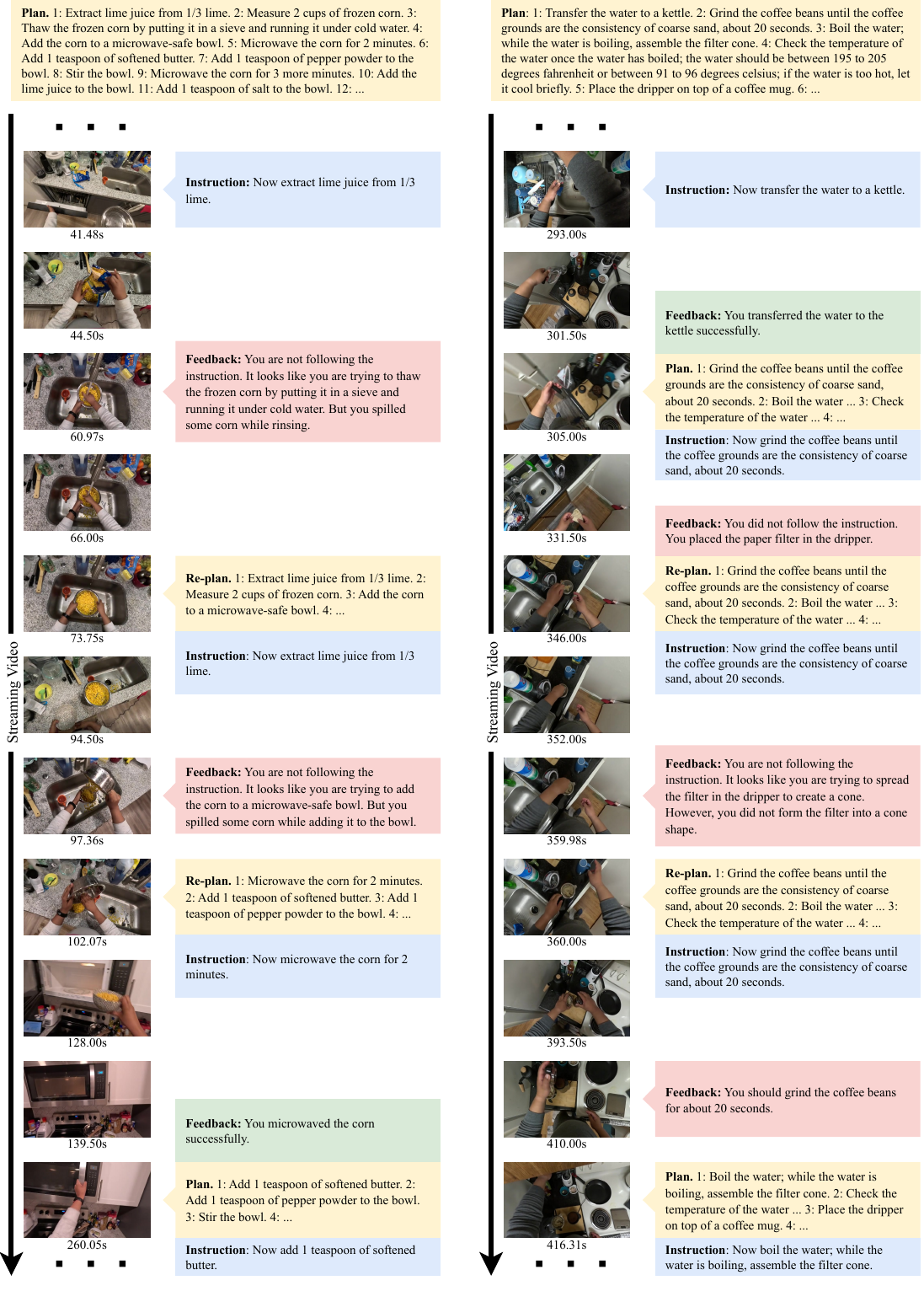}
    \caption{Data samples from the advanced planning set. Left: the user is preparing butter corn cup. Right: the user is making coffee.}
    \label{fig:advanced_data_2}
\end{figure}

In the advanced planning set (\cref{fig:advanced_data_1,fig:advanced_data_2}), we include cases where 
the user performs an action diverging from the instruction. Given a video with its associated graph-structured recipe, we append the actions that are not performed in the video (missing steps) to the ordered actions from the main set. We then topologically sort the actions using the Kahn's algorithm~\cite{kahn1962topological}. 
We discard some videos which can not be topologically sorted, because there exist actions in the video but absent from the recipe graph.
This (newly) sorted action order is used to build the step-by-step plan. Normally one action description forms one step in the plan. We group consecutive action descriptions into the same step only if there exists a step in the main set's plan containing the exact same action descriptions. 
Note that the action groups remain consistent with the main set.
An important difference to the main set is that, it is possible to have divergence between a step in the plan and the corresponding action group, in case the user does not follow the given instruction.

As in the main set, once the user finishes an action, we remove its description from the plan and remove any empty step; and once all the actions in the current action group are finished, we provide the instruction for the next step.
If the action group matches the corresponding step, that is, the user performs actions following the given instruction, we follow the same convention as in the main set. 
If the user performs an action diverging from the given instruction, we notify the user in the feedback. Specifically, if the action is performed without mistakes but does not follow the instruction, the feedback is given at the action's end time, starting with the sentence ``{\tt You did not follow the instruction.}'' and describing the action that is performed. If the action performed contains mistakes, the feedback is given when the mistake occurs, starting with the sentence ``{\tt You are not following the instruction.}'', describing the action that the user is trying to perform based on the initial plan, and pointing out the mistake.

In cases where the current step is not empty after all actions in the current action group finish, that is, the instruction is not followed at all or partially completed, we manually decide whether to keep or remove the current step in the plan based on the actions that the user has performed. 
This corresponds to the cases that require re-planning as the set 
of future steps need to be updated. 
\cref{tab:replan_stats} shows the ``re-plan'' statistics in our advanced planning set. As can be seen from the table, re-planning is required for about every 2.7 instructions or every 2.6 minutes on average.
\cref{fig:advanced_data_1,fig:advanced_data_2} show data samples from the advanced planning set. Note, in \cref{fig:advanced_data_1,fig:advanced_data_2}, ``Plan'' means that the action follows the instruction, and thus once the current step is finished, it is removed from the step-by-step plan. ``Re-plan'' means that the action diverges from the instruction, and when giving the next instruction, we need to decide whether to (re)instruct the user about the current step or not.

\section{\model{} Re-planner}
The re-planner is invoked (during inference) whenever the user does not follow the given instruction. This only occurs in the advanced planning set of the \benchmark{} benchmark. To identify such cases, we train the \model{} to add a ``{\tt You are not following the instruction.}'' or ``{\tt You did not follow the instruction.}'' prefix to feedback messages in case the user does the follow the given instruction. Note that, during training, the \model{} model is trained using the ground truth sequence of instructions (as the \model{} model is trained on a single recipe step at a time) and the re-planner is not used. During inference as we do have have the ground truth sequence of instructions, we use the re-planner to update the step-by-step instruction plan and provide the correct instructions (and feedbacks). We do this in two steps. 

First, the re-planner uses the Qwen3-32B~\cite{abs-2505-09388} model to extract the recipe step performed instead of the instructed recipe step using the following prompt:

\begin{tcolorbox}[
    colback=blue!5!white, 
    colframe=blue!75!black, 
    title=\textbf{Retrieve the performed action},
    fonttitle=\bfseries,
    coltitle=black, 
    colbacktitle=blue!20!white, 
    sharp corners,
    boxrule=0.5pt
]
You are an expert cooking instructor. You are observing a user cooking a given recipe step by step. 

\hrulefill 

\textbf{\#\#INSTRUCTIONS:}\\
Here are the recipe steps: [recipe\_steps].

The last instruction that you provided to the person is: [last\_instruction]. The person did not follow your instruction and performed a different recipe step by mistake which did not correspond to the provided instruction. So you provided this feedback to the person: [last\_feedback]. Which recipe step did the person likely perform instead of the step in the last instruction. RETURN THE RECIPE STEP AS A PYTHON STRING. ENSURE THAT YOU OUTPUT A RECIPE STEP AND DO NOT OUTPUT ANYTHING OTHER THAN A RECIPE STEP.
\end{tcolorbox}

Secondly, the re-planner determines if the current instruction needs to be repeated, based on the 
recipe steps and the past steps completed by the person. We use the following prompt for the Qwen3-32B~\cite{abs-2505-09388} model:

\begin{tcolorbox}[
    colback=blue!5!white, 
    colframe=blue!75!black, 
    title=\textbf{Decide whether repeating the last instruction},
    fonttitle=\bfseries,
    coltitle=black, 
    colbacktitle=blue!20!white, 
    sharp corners,
    boxrule=0.5pt
]
You are an expert cooking assistant. You are helping a user to make [recipe\_name], according to the following recipe steps: [recipe\_steps].

\hrulefill 

\textbf{\#\#INSTRUCTIONS:}\\
The user has already completed [past\_completed\_step\_counts] steps: [past\_completed\_steps].
Decide whether it is appropriate now to ask the user to [last\_instructed\_action], considering the effect of all the steps that the user performed.
Your answer must begin with 'Yes' or 'No', followed by an explanation.
\end{tcolorbox}

If the answer is ``{\tt No}'', that is, the last instruction is not to be repeated, we find all parent actions of the last action that the user performed from the corresponding recipe graph, and remove those actions from the step-by-step plan if they exist. Otherwise, we do not further update the step-by-step plan.

\section{\model{}: Training Details}
\subsection{Pre-training Data} As discussed in the main paper, for object grounding, \model{} is trained on image datasets, \ie, LVIS \cite{GuptaDG19} (for diverse household objects) and on video data using VISOR annotations from EPIC-KITCHENS \cite{Damen2022RESCALING}. The data follows a question answer format, where the questions are of the following format (following \cite{LiuLWL23a}):

\begin{tcolorbox}[
    colback=gray!5!white, 
    colframe=gray!75!black, 
    title=\textbf{LVIS / EPIC-KITCHENS: Grounding Questions},
    fonttitle=\bfseries,
    coltitle=black, 
    colbacktitle=gray!20!white, 
    sharp corners,
    boxrule=0.5pt
]

\begin{itemize}
    \item  Please provide the bounding box coordinates for the \rule{1cm}{0.15mm}., A: \rule{1cm}{0.15mm} 
    \item  Where is the \rule{1cm}{0.15mm} located in the image?,A: It is located at \rule{1cm}{0.15mm}.
    \item  What are the coordinates of the bounding box encompassing the \rule{1cm}{0.15mm}?, A: \rule{1cm}{0.15mm}.
    \item Can you pinpoint the \rule{1cm}{0.15mm} in the image by giving me its bounding box coordinates?, A: Yes, it is at \rule{1cm}{0.15mm}.
    \item Where is the \rule{1cm}{0.15mm} in the image? Give me the bounding box coordinates., A: \rule{1cm}{0.15mm}.
    \item I need the top-left and bottom-right corners of the <1>'s bounding box. What are they?, A: \rule{1cm}{0.15mm}.
    \item Could you identify the \rule{1cm}{0.15mm} in the image and tell me its location using bounding box coordinates?, A: Yes, it is at \rule{1cm}{0.15mm}.
\end{itemize}

\end{tcolorbox}

\begin{tcolorbox}[
    colback=gray!5!white, 
    colframe=gray!75!black, 
    sharp corners,
    boxrule=0.5pt
]

\begin{itemize}
    \item Describe the image concisely., A: \rule{1cm}{0.15mm}.
    \item Provide a brief description of the given image., A: \rule{1cm}{0.15mm}.
    \item Offer a succinct explanation of the picture presented., A: \rule{1cm}{0.15mm}.
    \item Summarize the visual content of the image., A: \rule{1cm}{0.15mm}.
    \item Give a short and clear explanation of the subsequent image., A: \rule{1cm}{0.15mm}.
    \item Share a concise interpretation of the image provided., A; \rule{1cm}{0.15mm}.
    \item Present a compact description of the photo’s key features., A: \rule{1cm}{0.15mm}.
    \item Relay a brief, clear account of the picture shown., A; \rule{1cm}{0.15mm}.
    \item Render a clear and concise summary of the photo., A: \rule{1cm}{0.15mm}.
    \item Write a terse but informative summary of the picture., A: \rule{1cm}{0.15mm}.
    \item Create a compact narrative representing the image presented., A: \rule{1cm}{0.15mm}.
\end{itemize}

\end{tcolorbox}

For action recognition tasks on SSv2~\cite{Goyal_2017_ICCV}, we use the following format:

\begin{tcolorbox}[
    colback=gray!5!white, 
    colframe=gray!75!black, 
    title=\textbf{SSv2: Action Recognition Questions},
    fonttitle=\bfseries,
    coltitle=black, 
    colbacktitle=gray!20!white, 
    sharp corners,
    boxrule=0.5pt
]

\begin{itemize}
    \item  Describe the action I performed in this video in detail and name the objects that the I interacts with?, A: \rule{1cm}{0.15mm}.
    \item Can you provide a step-by-step description of the action in the video which includes the specific objects that I touched or manipulated?, A: \rule{1cm}{0.15mm}.
    \item Tell me a description of  the action performed by me in the video that includes the names of any items that I interact with., A: \rule{1cm}{0.15mm}.
    \item What action happens in the video and what objects are involved in the action?, A: \rule{1cm}{0.15mm}.
    \item Describe my actions and the objects that I come across., A: \rule{1cm}{0.15mm}.
    
\end{itemize}

\end{tcolorbox}

For narration tasks on EPIC-KITCHENS we convert the action descriptions to second person format, \eg, ``pick up plate'' to ``You picked up a plate.'' For Ego4D, we employ the narrations used by \cite{VideoLLM-online}.

\subsection{Fine-tuning Data} As mentioned in the main paper, EPIC-KITCHENS and Ego4D datasets included in the fine-tuning phase do not include instances of mistakes. Therefore, we formulate a novel data augmentation scheme to generate (counterfactual) mistakes. First, we describe the process used on Ego4D to generate counterfactual action descriptions in detail. 

\myparagraph{Counterfactual Mistake Augmentation.} To this end, we utilize the Ego4D goalstep annotations \cite{SongBNWMT23} and the FHO annotations \cite{Grauman_2022_CVPR}. The FHO annotations largely include fine-grained short duration actions while the goalstep annotations include longer-ranged actions more closely aligned with the recipe steps in our \benchmark{} benchmark. To create temporally localized counterfactual mistakes, we first find the FHO actions that are included within each goalstep action. Then we ask Qwen-2.5-32B-Instruct~\cite{abs-2412-15115} identify the ``critical'' FHO action for each goalstep action, \eg, for the goalstep action \texttt{wash green beans in water}, the critical FHO identified by Qwen-2.5-32B-Instruct is \texttt{washes green bean} (from the following FHO actions within the goalstep action: \texttt{collects green bean}, \texttt{puts green beans on the chopping board}, \texttt{puts green beans in cooking pan}, $\dots$, \texttt{picks green beans}, \texttt{opens tap}, \texttt{washes green bean}, $\dots$). We then ask the Qwen-2.5-32B-Instruct to construct two counterfactual actions. Firstly, by changing the noun to an alternative noun in the scene, \ie, \texttt{washes green bean} to \texttt{washes carrots}. The list of nouns in the scene is generated by identifying objects in the scene using DETR \cite{CarionMSUKZ20}. Secondly, by proposing an alternative verb, \ie, \texttt{washes green bean} to \texttt{mash green bean}. Then we use this counterfactual action to create an instruction and feedback pair. We use the point-of-no-return timestamp \cite{Grauman_2022_CVPR} for the mistake feedback.

In case of the EPIC-KITCHENS dataset, most actions are of short duration (<10 seconds long). We directly use Qwen-2.5-32B-Instruct to construct two counterfactual actions by using an alternative noun in the scene and by an alternative verb as described above. As the actions are short we use the end of action timestamp to generate the (mistake) feedback.

\subsection{Training and Inference Hyperparameters.} We use input video resolution of $448\times448$ at 2 fps. The InternViT-300M-448px-V2\_5 vision head produces $N=1025$ tokens (including the CLS token) per input frame. 
We use the mechanism outlined in VisionZip \cite{YangCTWL0J25} to reduce the number of tokens to 256. Then, our Q-Former reduces this further to $K=32$ tokens. 

In addition to the vision features, the Mamba-130M language backbone of the \model{} model is prompted with the following prompt:
\begin{tcolorbox}[
    colback=blue!5!white, 
    colframe=blue!75!black, 
    title=\textbf{\model{}: Interactive Inference (Language Backbone)},
    fonttitle=\bfseries,
    coltitle=black, 
    colbacktitle=blue!20!white, 
    sharp corners,
    boxrule=0.5pt
]

You are an expert cooking assistant that is helping a person cook the following step by step recipe: [recipe\_steps].\\
You just provided the following instruction to the person: [last\_instruction]. \\
Now watch the video and provide the appropriate success or failure messages.
\end{tcolorbox}

The \model{} model is trained using 8 Nvidia H100 GPUs. We use the AdamW~\cite{LoshchilovH19} optimizer. During the pre-training phase, we train only the Q-Former and the \model{} model is trained using a learning rate of $1\times10^{-5}$ for ~200k iterations. 
We again use a learning rate of $1\times10^{-5}$, for ~120k iterations. 
During the fine-tuning phase, we train on single recipe steps and clip the maximum length to 3 minutes. 
During inference, we re-initialize the \model{} model after every recipe step.

\section{\model{} Results}
\begin{figure}[h!] 
    \centering
    \begin{subfigure}[b]{\linewidth} 
        \centering
        \includegraphics[width=\linewidth]{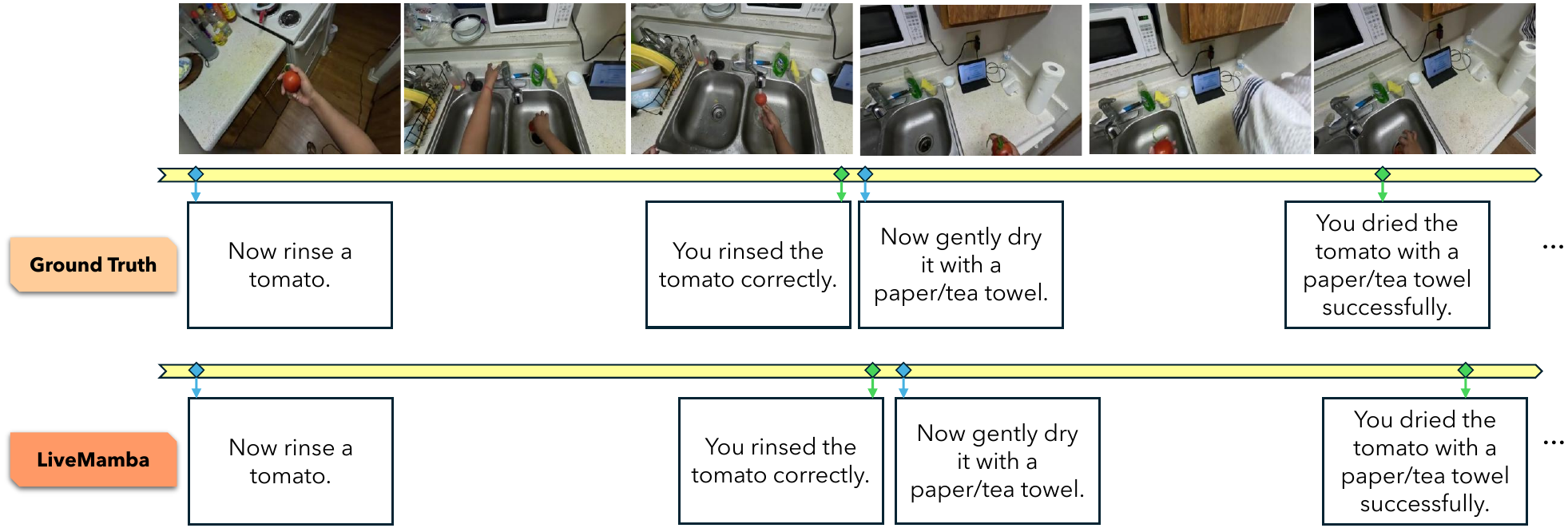} 
    \end{subfigure}
    \par\bigskip
    \begin{subfigure}[b]{\linewidth}
        \centering
        \includegraphics[width=\linewidth]{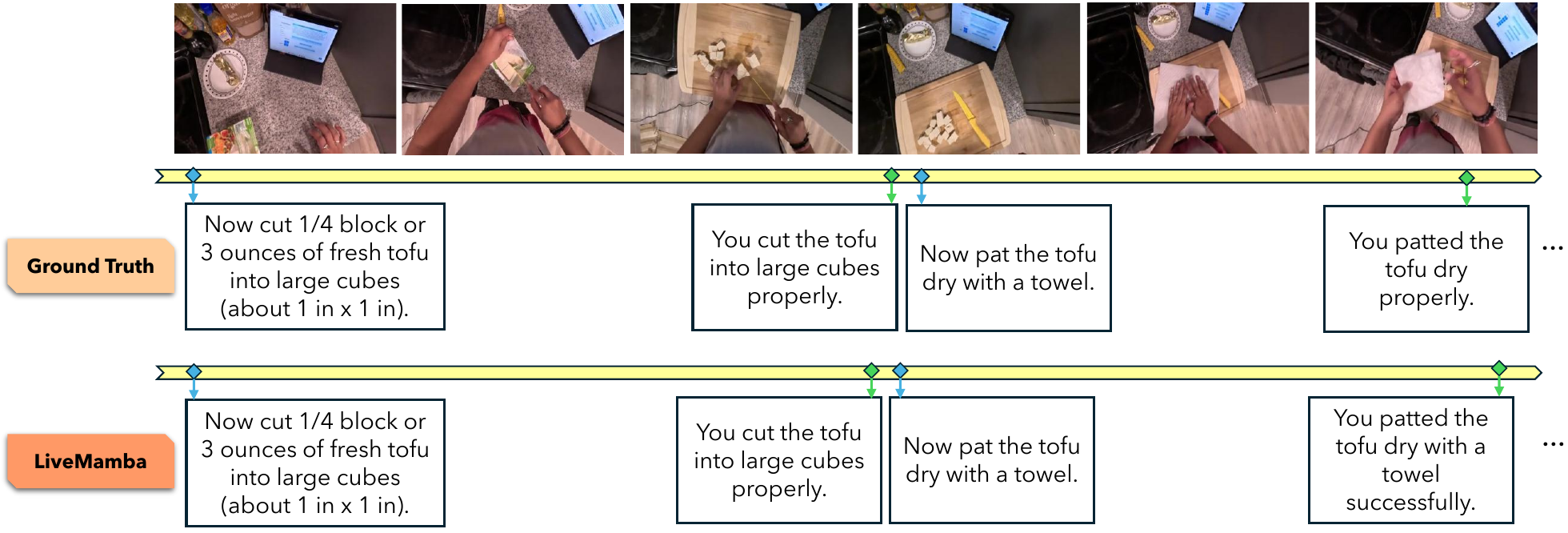}
    \end{subfigure}
    \par\bigskip
    \begin{subfigure}[b]{\linewidth}
        \centering
        \includegraphics[width=\linewidth]{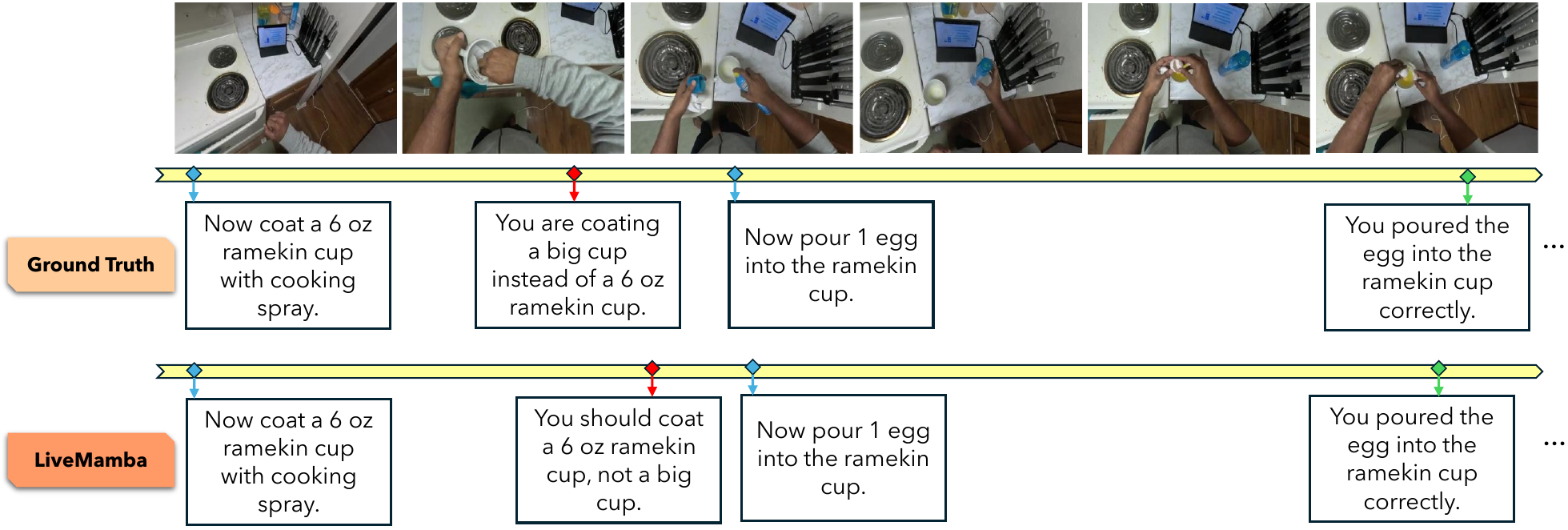}
    \end{subfigure}
    \caption{Predictions from our \model{} from the main set of the \benchmark{} benchmark.}
    \label{fig:livemamba_qual_supp} 
\end{figure}

In \cref{fig:livemamba_qual_supp}, we show additional qualitative examples of predictions by our \model{} model from the main set of the \benchmark{} benchmark. In the top two rows we see that our \model{} model is able to successfully recognize that the user has completed complex steps from the step-by-step plans, \eg, rinsing a tomato and cutting tofu into large cubes. In the third row we show an example where the user incorrectly coats a big cup instead of a 6-oz. cup with cooking spray. Our \model{} model is able to provide timely feedback in this case.

\section{Zero-shot Evaluation}
Here we provide the prompts used for zero-shot evaluation. We begin with a more detailed description of the prompting strategy in Section 5.2 of the main paper.

\myparagraph{Promoting Strategy.} This strategy involves prompting the model at regular intervals to detect both successful completions of instructions and mistakes. However prompting after every input frame is not computationally feasible. To balance accuracy and compute requirements, we prompt the models at an interval of 5 seconds. First, the model is asked if the user has completed the current instruction (obtained from the recipes in the \benchmark{} benchmark). We use the following prompt for the Gemini-2.5-Flash \cite{team2024gemini}, Qwen2.5-VL-7B-Instruct \cite{abs-2502-13923}, Qwen2-VL-7B-Instruct \cite{Qwen2VL}, VideoLLaMA3-7B~\cite{damonlpsg2025videollama3} models:

\begin{tcolorbox}[
    colback=blue!5!white, 
    colframe=blue!75!black, 
    title=\textbf{Gemini-2.5-Flash/ Qwen2x-VL-7B-Instruct / VideoLLaMA3-7B: Check if instruction complete},
    fonttitle=\bfseries,
    coltitle=black, 
    colbacktitle=blue!20!white, 
    sharp corners,
    boxrule=0.5pt
]

You are an expert cooking assistant helping a person cook. The person is provided with an instruction and your task is to check if the instruction has been completed.

\hrulefill 

\textbf{\#\#INSTRUCTIONS:}\\
The person has been instructed to: [instruction]. \\
If the person has completed the instruction answer ``yes'' else answer ``no''. DO NOT OUTPUT ANY OTHER TEXT.
\end{tcolorbox}

In case of VideoChat2~\cite{li2024mvbench} model, we use the following prompt:

\begin{tcolorbox}[
    colback=blue!5!white, 
    colframe=blue!75!black, 
    title=\textbf{VideoChat2: Check if instruction complete},
    fonttitle=\bfseries,
    coltitle=black, 
    colbacktitle=blue!20!white, 
    sharp corners,
    boxrule=0.5pt
]

You are an expert cooking assistant helping a person cook. The person is provided with an instruction and your task is to check if the instruction has been completed. You must check the video content very closely and confirm if the instruction has been completely followed and finished.

\hrulefill 

\textbf{\#\#INSTRUCTIONS:}\\
The person has been instructed to: [instruction]. \\
If the person has completed the instruction answer ``yes'' else answer ``no''. Answer ``yes'' ONLY IF the instructed action is completed in the video, otherwise answer ``no''.
\end{tcolorbox}

In case of the Video-LLaVA~\cite{zhu2023languagebind, lin2023video}, Video-ChatGPT~\cite{maaz-etal-2024-video}, LLaVA-NeXT~\cite{liu2024llavanext} as they do not accept system messages we use the following simplified prompt:

\begin{tcolorbox}[
    colback=blue!5!white, 
    colframe=blue!75!black, 
    title=\textbf{Video-LLaVA / Video-ChatGPT / LLaVA-NeXT : Check if instruction complete},
    fonttitle=\bfseries,
    coltitle=black, 
    colbacktitle=blue!20!white, 
    sharp corners,
    boxrule=0.5pt
]
The person has been instructed to: [instruction]. \\
If the person has completed the instruction answer ``yes'' else answer ``no''.
\end{tcolorbox}

If the model answers ``yes'', then we move on to the next instruction in the recipe. If not, then we ask the model to check for mistakes. 

For VideoLLM-Online -- since the model is trained to narrate videos, it can not detect completion and mistakes out of the box. Therefore, we use a helper LLM (Phi-3-mini-4k-Instruct) to help detect completion and mistakes given narrations. We use the following prompts,
\begin{tcolorbox}[
    colback=blue!5!white, 
    colframe=blue!75!black, 
    title=\textbf{VideoLLM-Online: Check if instruction is complete},
    fonttitle=\bfseries,
    coltitle=black, 
    colbacktitle=blue!20!white, 
    sharp corners,
    boxrule=0.5pt
]
\textbf{VideoLLM-Online:} Please narrate the video in real-time.

\textbf{Helper-LLM (Phi-3-mini-4k-Instruct):} You are an intelligent chatbot that is judging another system which narrates human cooking videos. Given a high level action instruction and a list of narrations generated from the system, your job is to decide if the narration is correct and shows completion of the instruction. Answer 'yes' if the instruction is completed otherwise output 'no'.

Instruction: [instruction]

Narrations: [videollm-online narrations]
\end{tcolorbox}

In the offline (turn-based) evaluation scheme used in CaptainCook4D \cite{PeddiACPVGZWKRR24}, multi-modal LLMs are prompted with sequence of questions per error category to detect mistakes. However, such a scheme is infeasible in case of streaming setups due to high computational costs. Therefore, to enable mistake detection in a single inference step, we design a prompt that concisely explains the possible mistakes that are possible while following a given instruction. The model can then use this information to recognize possible mistakes. In detail, for a given instruction we find similar instructions in the training set and use Qwen-2.5-32B-Instruct \cite{abs-2412-15115} to summarize the possible mistakes. The following are some examples:

\begin{tcolorbox}[
    colback=yellow!5!white, 
    colframe=black, 
    title=\textbf{Instruction: Now place 8-inch flour tortilla on cutting board.},
    fonttitle=\bfseries,
    coltitle=black, 
    colbacktitle=orange!20!white, 
    sharp corners,
    boxrule=0.5pt
]
When guiding the user through this step, the cooking instructor should watch out for these potential mistakes:
\begin{enumerate}
    \item Users might toast or heat the tortilla before placing it on the cutting board.
    \item Users may place the tortilla on a plate instead of a cutting board.
    \item Users could use an unclean surface instead of a clean cutting board.
\end{enumerate}
\end{tcolorbox}

\begin{tcolorbox}[
    colback=yellow!5!white, 
    colframe=black, 
    title=\textbf{Instruction: Now add 1/4 tsp salt to a bowl.},
    fonttitle=\bfseries,
    coltitle=black, 
    colbacktitle=orange!20!white, 
    sharp corners,
    boxrule=0.5pt
]
When guiding the user through this step, the cooking instructor should watch out for these potential mistakes:
\begin{enumerate}
    \item Spilling salt while measuring or adding it.
    \item Adding too much salt, specifically 1/2 tsp instead of the required 1/4 tsp.
    \item Confusing 1/3 tablespoon with 1/3 teaspoon.
    \item Accidentally adding the salt to the pan rather than the bowl.
\end{enumerate}
\end{tcolorbox}

We then use these ``mistake summaries'' to prompt the multi-modal LLM. For Gemini-2.5-Flash \cite{team2024gemini}, Qwen2.5-VL-7B-Instruct \cite{abs-2502-13923}, Qwen2-VL-7B-Instruct \cite{Qwen2VL}, VideoLLaMA3-7B~\cite{damonlpsg2025videollama3}, VideoChat2~\cite{li2024mvbench} we use the following prompts:

\begin{tcolorbox}[
    colback=blue!5!white, 
    colframe=blue!75!black, 
    title=\textbf{Gemini-2.5-Flash / Qwen2x-VL-7B-Instruct / VideoLLaMA3-7B/ VideoChat2: Check if instruction complete},
    fonttitle=\bfseries,
    coltitle=black, 
    colbacktitle=blue!20!white, 
    sharp corners,
    boxrule=0.5pt
]

You are an expert cooking assistant who is observing a person who is provided with step by step instructions for cooking. You should look out for mistakes made by the person.

\hrulefill 

\textbf{\#\#INSTRUCTIONS:}\\
The person is trying to complete the following instruction: [instruction]. \\
This is how you can check for mistakes: [mistake summary]. \\
Your task is to check if the person has already made a mistake. \\
Note that the person may not have completed the provided instruction, that is, the person may have only partially completed the provided instruction.\\
The answer should be ``yes'' or ``no''. In case of yes, please provide a concise feedback to the person describing the mistake (i.e. Yes. <feedback>.). Directly address the person.
\end{tcolorbox}

In case of Video-LLaVA~\cite{zhu2023languagebind, lin2023video}, Video-ChatGPT~\cite{maaz-etal-2024-video}, LLaVA-NeXT~\cite{liu2024llavanext} models, they do not support complex prompts as they were mainly designed for question answering tasks. They we employ an alternative strategy where we ask the model to narrate the video in 30 second chunks. Then we use the (concatenated) past narrations since the last instruction completion to prompt the model to look for mistakes.

\begin{tcolorbox}[
    colback=blue!5!white, 
    colframe=blue!75!black, 
    title=\textbf{Video-LLaVA / Video-ChatGPT / LLaVA-NeXT : Check if instruction complete},
    fonttitle=\bfseries,
    coltitle=black, 
    colbacktitle=blue!20!white, 
    sharp corners,
    boxrule=0.5pt
]
The person has been instructed to: [instruction]. \\
ill now the person has done the following: [past\_narrations].\\
Your task is to check if the person has made a mistake. \\
The answer should be ``yes'' or ``no''. In case of ``yes'', please provide a feedback to the user describing the mistake. Directly address the person.\\
\end{tcolorbox}

For VideoLLM-Online, we use a similar strategy as done in completion detection. We feed VideoLLM-Online's narration to a helper LLM to get mistake detection. We use the following prompts,
\begin{tcolorbox}[
    colback=blue!5!white, 
    colframe=blue!75!black, 
    title=\textbf{VideoLLM-Online: Check if instruction is complete},
    fonttitle=\bfseries,
    coltitle=black, 
    colbacktitle=blue!20!white, 
    sharp corners,
    boxrule=0.5pt
]
\textbf{VideoLLM-Online:} Please narrate the video in real-time.

\textbf{Helper-LLM (Phi-3-mini-4k-Instruct):} You are an intelligent chatbot that is judging another system which narrates human cooking videos. Given a high level action instruction and a list of narrations generated from the system, your job is to decide if a mistake has been made. 

The user has be instructed to do the following: [instruction] \\
Till now the person has done the following: [videollm-online narrations] \\
Your task is to check if the person has made a mistake. \\
The answer should be yes or no. In case of yes, please provide a feedback to the user describing the mistake. Directly address the person.
\end{tcolorbox}

\FloatBarrier
\newpage
\section*{NeurIPS Paper Checklist}

\begin{enumerate}

\item {\bf Claims}
    \item[] Question: Do the main claims made in the abstract and introduction accurately reflect the paper's contributions and scope?
    \item[] Answer: \answerYes{} 
    \item[] Justification: The abstract and introduction clearly states that the main contributions of the paper are a novel method: \model{} and a dataset: \benchmark{}.
    \item[] Guidelines:
    \begin{itemize}
        \item The answer NA means that the abstract and introduction do not include the claims made in the paper.
        \item The abstract and/or introduction should clearly state the claims made, including the contributions made in the paper and important assumptions and limitations. A No or NA answer to this question will not be perceived well by the reviewers. 
        \item The claims made should match theoretical and experimental results, and reflect how much the results can be expected to generalize to other settings. 
        \item It is fine to include aspirational goals as motivation as long as it is clear that these goals are not attained by the paper. 
    \end{itemize}

\item {\bf Limitations}
    \item[] Question: Does the paper discuss the limitations of the work performed by the authors?
    \item[] Answer: \answerYes{} 
    \item[] Justification: We discuss the limitations in the ``Limitations'' section.
    \item[] Guidelines:
    \begin{itemize}
        \item The answer NA means that the paper has no limitation while the answer No means that the paper has limitations, but those are not discussed in the paper. 
        \item The authors are encouraged to create a separate "Limitations" section in their paper.
        \item The paper should point out any strong assumptions and how robust the results are to violations of these assumptions (e.g., independence assumptions, noiseless settings, model well-specification, asymptotic approximations only holding locally). The authors should reflect on how these assumptions might be violated in practice and what the implications would be.
        \item The authors should reflect on the scope of the claims made, e.g., if the approach was only tested on a few datasets or with a few runs. In general, empirical results often depend on implicit assumptions, which should be articulated.
        \item The authors should reflect on the factors that influence the performance of the approach. For example, a facial recognition algorithm may perform poorly when image resolution is low or images are taken in low lighting. Or a speech-to-text system might not be used reliably to provide closed captions for online lectures because it fails to handle technical jargon.
        \item The authors should discuss the computational efficiency of the proposed algorithms and how they scale with dataset size.
        \item If applicable, the authors should discuss possible limitations of their approach to address problems of privacy and fairness.
        \item While the authors might fear that complete honesty about limitations might be used by reviewers as grounds for rejection, a worse outcome might be that reviewers discover limitations that aren't acknowledged in the paper. The authors should use their best judgment and recognize that individual actions in favor of transparency play an important role in developing norms that preserve the integrity of the community. Reviewers will be specifically instructed to not penalize honesty concerning limitations.
    \end{itemize}

\item {\bf Theory assumptions and proofs}
    \item[] Question: For each theoretical result, does the paper provide the full set of assumptions and a complete (and correct) proof?
    \item[] Answer: \answerNA{} 
    \item[] Justification: There are no theoretical results.
    \item[] Guidelines:
    \begin{itemize}
        \item The answer NA means that the paper does not include theoretical results. 
        \item All the theorems, formulas, and proofs in the paper should be numbered and cross-referenced.
        \item All assumptions should be clearly stated or referenced in the statement of any theorems.
        \item The proofs can either appear in the main paper or the supplemental material, but if they appear in the supplemental material, the authors are encouraged to provide a short proof sketch to provide intuition. 
        \item Inversely, any informal proof provided in the core of the paper should be complemented by formal proofs provided in appendix or supplemental material.
        \item Theorems and Lemmas that the proof relies upon should be properly referenced. 
    \end{itemize}

    \item {\bf Experimental result reproducibility}
    \item[] Question: Does the paper fully disclose all the information needed to reproduce the main experimental results of the paper to the extent that it affects the main claims and/or conclusions of the paper (regardless of whether the code and data are provided or not)?
    \item[] Answer: \answerYes{} 
    \item[] Justification: Yes, experimental details are included in the main paper and appendix. 
    \item[] Guidelines:
    \begin{itemize}
        \item The answer NA means that the paper does not include experiments.
        \item If the paper includes experiments, a No answer to this question will not be perceived well by the reviewers: Making the paper reproducible is important, regardless of whether the code and data are provided or not.
        \item If the contribution is a dataset and/or model, the authors should describe the steps taken to make their results reproducible or verifiable. 
        \item Depending on the contribution, reproducibility can be accomplished in various ways. For example, if the contribution is a novel architecture, describing the architecture fully might suffice, or if the contribution is a specific model and empirical evaluation, it may be necessary to either make it possible for others to replicate the model with the same dataset, or provide access to the model. In general. releasing code and data is often one good way to accomplish this, but reproducibility can also be provided via detailed instructions for how to replicate the results, access to a hosted model (e.g., in the case of a large language model), releasing of a model checkpoint, or other means that are appropriate to the research performed.
        \item While NeurIPS does not require releasing code, the conference does require all submissions to provide some reasonable avenue for reproducibility, which may depend on the nature of the contribution. For example
        \begin{enumerate}
            \item If the contribution is primarily a new algorithm, the paper should make it clear how to reproduce that algorithm.
            \item If the contribution is primarily a new model architecture, the paper should describe the architecture clearly and fully.
            \item If the contribution is a new model (e.g., a large language model), then there should either be a way to access this model for reproducing the results or a way to reproduce the model (e.g., with an open-source dataset or instructions for how to construct the dataset).
            \item We recognize that reproducibility may be tricky in some cases, in which case authors are welcome to describe the particular way they provide for reproducibility. In the case of closed-source models, it may be that access to the model is limited in some way (e.g., to registered users), but it should be possible for other researchers to have some path to reproducing or verifying the results.
        \end{enumerate}
    \end{itemize}

\item {\bf Open access to data and code}
    \item[] Question: Does the paper provide open access to the data and code, with sufficient instructions to faithfully reproduce the main experimental results, as described in supplemental material?
    \item[] Answer: \answerNo{} 
    \item[] Justification: The \benchmark{} dataset released at \url{https://huggingface.co/qualcomm/qualcomm-interactive-cooking-dataset}.
    \item[] Guidelines:
    \begin{itemize}
        \item The answer NA means that paper does not include experiments requiring code.
        \item Please see the NeurIPS code and data submission guidelines (\url{https://nips.cc/public/guides/CodeSubmissionPolicy}) for more details.
        \item While we encourage the release of code and data, we understand that this might not be possible, so “No” is an acceptable answer. Papers cannot be rejected simply for not including code, unless this is central to the contribution (e.g., for a new open-source benchmark).
        \item The instructions should contain the exact command and environment needed to run to reproduce the results. See the NeurIPS code and data submission guidelines (\url{https://nips.cc/public/guides/CodeSubmissionPolicy}) for more details.
        \item The authors should provide instructions on data access and preparation, including how to access the raw data, preprocessed data, intermediate data, and generated data, etc.
        \item The authors should provide scripts to reproduce all experimental results for the new proposed method and baselines. If only a subset of experiments are reproducible, they should state which ones are omitted from the script and why.
        \item At submission time, to preserve anonymity, the authors should release anonymized versions (if applicable).
        \item Providing as much information as possible in supplemental material (appended to the paper) is recommended, but including URLs to data and code is permitted.
    \end{itemize}

\item {\bf Experimental setting/details}
    \item[] Question: Does the paper specify all the training and test details (e.g., data splits, hyperparameters, how they were chosen, type of optimizer, etc.) necessary to understand the results?
    \item[] Answer: \answerYes{} 
    \item[] Justification: The details are provided in the main paper and the appendix.
    \item[] Guidelines:
    \begin{itemize}
        \item The answer NA means that the paper does not include experiments.
        \item The experimental setting should be presented in the core of the paper to a level of detail that is necessary to appreciate the results and make sense of them.
        \item The full details can be provided either with the code, in appendix, or as supplemental material.
    \end{itemize}

\item {\bf Experiment statistical significance}
    \item[] Question: Does the paper report error bars suitably and correctly defined or other appropriate information about the statistical significance of the experiments?
    \item[] Answer: \answerNA{} 
    \item[] Justification: While a full statistical analysis of the performance of multi-modal LLMs are computationally prohibitive, our initial analysis suggest limited variance of results on the \benchmark{} benchmark.
    \item[] Guidelines:
    \begin{itemize}
        \item The answer NA means that the paper does not include experiments.
        \item The authors should answer "Yes" if the results are accompanied by error bars, confidence intervals, or statistical significance tests, at least for the experiments that support the main claims of the paper.
        \item The factors of variability that the error bars are capturing should be clearly stated (for example, train/test split, initialization, random drawing of some parameter, or overall run with given experimental conditions).
        \item The method for calculating the error bars should be explained (closed form formula, call to a library function, bootstrap, etc.)
        \item The assumptions made should be given (e.g., Normally distributed errors).
        \item It should be clear whether the error bar is the standard deviation or the standard error of the mean.
        \item It is OK to report 1-sigma error bars, but one should state it. The authors should preferably report a 2-sigma error bar than state that they have a 96\% CI, if the hypothesis of Normality of errors is not verified.
        \item For asymmetric distributions, the authors should be careful not to show in tables or figures symmetric error bars that would yield results that are out of range (e.g. negative error rates).
        \item If error bars are reported in tables or plots, The authors should explain in the text how they were calculated and reference the corresponding figures or tables in the text.
    \end{itemize}

\item {\bf Experiments compute resources}
    \item[] Question: For each experiment, does the paper provide sufficient information on the computer resources (type of compute workers, memory, time of execution) needed to reproduce the experiments?
    \item[] Answer: \answerYes{} 
    \item[] Justification: Yes, the details are provided in the appendix.
    \item[] Guidelines:
    \begin{itemize}
        \item The answer NA means that the paper does not include experiments.
        \item The paper should indicate the type of compute workers CPU or GPU, internal cluster, or cloud provider, including relevant memory and storage.
        \item The paper should provide the amount of compute required for each of the individual experimental runs as well as estimate the total compute. 
        \item The paper should disclose whether the full research project required more compute than the experiments reported in the paper (e.g., preliminary or failed experiments that didn't make it into the paper). 
    \end{itemize}
    
\item {\bf Code of ethics}
    \item[] Question: Does the research conducted in the paper conform, in every respect, with the NeurIPS Code of Ethics \url{https://neurips.cc/public/EthicsGuidelines}?
    \item[] Answer: \answerYes{} 
    \item[] Justification: Yes.
    \item[] Guidelines:
    \begin{itemize}
        \item The answer NA means that the authors have not reviewed the NeurIPS Code of Ethics.
        \item If the authors answer No, they should explain the special circumstances that require a deviation from the Code of Ethics.
        \item The authors should make sure to preserve anonymity (e.g., if there is a special consideration due to laws or regulations in their jurisdiction).
    \end{itemize}

\item {\bf Broader impacts}
    \item[] Question: Does the paper discuss both potential positive societal impacts and negative societal impacts of the work performed?
    \item[] Answer: \answerYes{} 
    \item[] Justification: We include a "Broader Impacts" section.
    \item[] Guidelines:
    \begin{itemize}
        \item The answer NA means that there is no societal impact of the work performed.
        \item If the authors answer NA or No, they should explain why their work has no societal impact or why the paper does not address societal impact.
        \item Examples of negative societal impacts include potential malicious or unintended uses (e.g., disinformation, generating fake profiles, surveillance), fairness considerations (e.g., deployment of technologies that could make decisions that unfairly impact specific groups), privacy considerations, and security considerations.
        \item The conference expects that many papers will be foundational research and not tied to particular applications, let alone deployments. However, if there is a direct path to any negative applications, the authors should point it out. For example, it is legitimate to point out that an improvement in the quality of generative models could be used to generate deepfakes for disinformation. On the other hand, it is not needed to point out that a generic algorithm for optimizing neural networks could enable people to train models that generate Deepfakes faster.
        \item The authors should consider possible harms that could arise when the technology is being used as intended and functioning correctly, harms that could arise when the technology is being used as intended but gives incorrect results, and harms following from (intentional or unintentional) misuse of the technology.
        \item If there are negative societal impacts, the authors could also discuss possible mitigation strategies (e.g., gated release of models, providing defenses in addition to attacks, mechanisms for monitoring misuse, mechanisms to monitor how a system learns from feedback over time, improving the efficiency and accessibility of ML).
    \end{itemize}
    
\item {\bf Safeguards}
    \item[] Question: Does the paper describe safeguards that have been put in place for responsible release of data or models that have a high risk for misuse (e.g., pretrained language models, image generators, or scraped datasets)?
    \item[] Answer: \answerNA{} 
    \item[] Justification: Our model and data is strictly applicable only to cooking scenarios from an ego-centric viewpoint.
    \item[] Guidelines:
    \begin{itemize}
        \item The answer NA means that the paper poses no such risks.
        \item Released models that have a high risk for misuse or dual-use should be released with necessary safeguards to allow for controlled use of the model, for example by requiring that users adhere to usage guidelines or restrictions to access the model or implementing safety filters. 
        \item Datasets that have been scraped from the Internet could pose safety risks. The authors should describe how they avoided releasing unsafe images.
        \item We recognize that providing effective safeguards is challenging, and many papers do not require this, but we encourage authors to take this into account and make a best faith effort.
    \end{itemize}

\item {\bf Licenses for existing assets}
    \item[] Question: Are the creators or original owners of assets (e.g., code, data, models), used in the paper, properly credited and are the license and terms of use explicitly mentioned and properly respected?
    \item[] Answer: \answerYes{} 
    \item[] Justification: All licenses are respected. 
    \item[] Guidelines:
    \begin{itemize}
        \item The answer NA means that the paper does not use existing assets.
        \item The authors should cite the original paper that produced the code package or dataset.
        \item The authors should state which version of the asset is used and, if possible, include a URL.
        \item The name of the license (e.g., CC-BY 4.0) should be included for each asset.
        \item For scraped data from a particular source (e.g., website), the copyright and terms of service of that source should be provided.
        \item If assets are released, the license, copyright information, and terms of use in the package should be provided. For popular datasets, \url{paperswithcode.com/datasets} has curated licenses for some datasets. Their licensing guide can help determine the license of a dataset.
        \item For existing datasets that are re-packaged, both the original license and the license of the derived asset (if it has changed) should be provided.
        \item If this information is not available online, the authors are encouraged to reach out to the asset's creators.
    \end{itemize}

\item {\bf New assets}
    \item[] Question: Are new assets introduced in the paper well documented and is the documentation provided alongside the assets?
    \item[] Answer: \answerYes{} 
    \item[] Justification: The \benchmark{} dataset released at \url{https://huggingface.co/qualcomm/qualcomm-interactive-cooking-dataset}.
    \item[] Guidelines:
    \begin{itemize}
        \item The answer NA means that the paper does not release new assets.
        \item Researchers should communicate the details of the dataset/code/model as part of their submissions via structured templates. This includes details about training, license, limitations, etc. 
        \item The paper should discuss whether and how consent was obtained from people whose asset is used.
        \item At submission time, remember to anonymize your assets (if applicable). You can either create an anonymized URL or include an anonymized zip file.
    \end{itemize}

\item {\bf Crowdsourcing and research with human subjects}
    \item[] Question: For crowdsourcing experiments and research with human subjects, does the paper include the full text of instructions given to participants and screenshots, if applicable, as well as details about compensation (if any)? 
    \item[] Answer: \answerNA{} 
    \item[] Justification: The \benchmark{} dataset and benchmark was not collected by crowd sourcing. The annotations were generated by the authors themselves.
    \item[] Guidelines:
    \begin{itemize}
        \item The answer NA means that the paper does not involve crowdsourcing nor research with human subjects.
        \item Including this information in the supplemental material is fine, but if the main contribution of the paper involves human subjects, then as much detail as possible should be included in the main paper. 
        \item According to the NeurIPS Code of Ethics, workers involved in data collection, curation, or other labor should be paid at least the minimum wage in the country of the data collector. 
    \end{itemize}

\item {\bf Institutional review board (IRB) approvals or equivalent for research with human subjects}
    \item[] Question: Does the paper describe potential risks incurred by study participants, whether such risks were disclosed to the subjects, and whether Institutional Review Board (IRB) approvals (or an equivalent approval/review based on the requirements of your country or institution) were obtained?
    \item[] Answer: \answerNA{} 
    \item[] Justification: The paper does not involve crowdsourcing nor research with human subjects.
    \item[] Guidelines:
    \begin{itemize}
        \item The answer NA means that the paper does not involve crowdsourcing nor research with human subjects.
        \item Depending on the country in which research is conducted, IRB approval (or equivalent) may be required for any human subjects research. If you obtained IRB approval, you should clearly state this in the paper. 
        \item We recognize that the procedures for this may vary significantly between institutions and locations, and we expect authors to adhere to the NeurIPS Code of Ethics and the guidelines for their institution. 
        \item For initial submissions, do not include any information that would break anonymity (if applicable), such as the institution conducting the review.
    \end{itemize}

\item {\bf Declaration of LLM usage}
    \item[] Question: Does the paper describe the usage of LLMs if it is an important, original, or non-standard component of the core methods in this research? Note that if the LLM is used only for writing, editing, or formatting purposes and does not impact the core methodology, scientific rigorousness, or originality of the research, declaration is not required.
    \item[] Answer: \answerYes{} 
    \item[] Justification: Yes, all uses of LLMs are clearly described.
    \item[] Guidelines:
    \begin{itemize}
        \item The answer NA means that the core method development in this research does not involve LLMs as any important, original, or non-standard components.
        \item Please refer to our LLM policy (\url{https://neurips.cc/Conferences/2025/LLM}) for what should or should not be described.
    \end{itemize}

\end{enumerate}

\end{document}